\documentclass{article}
\usepackage{spconf,amsmath,graphicx}
\usepackage{amsfonts}
\usepackage[table,xcdraw]{xcolor}
\usepackage[]{algorithm2e}
\usepackage{multirow}
\usepackage{placeins}
\usepackage{comment}
\usepackage{url}
\usepackage{cite}
\usepackage{pgfplots}
\usepackage{siunitx}
\pgfplotsset{compat=1.9}
\usepgfplotslibrary{statistics}
\usetikzlibrary{pgfplots.groupplots}
\usetikzlibrary{patterns,positioning,fadings,through}
\usepackage{amsfonts}
\usepackage{multirow}
\usepackage{siunitx}
\usepackage[normalem]{ulem}
\usepackage{adjustbox}
\usepackage{balance}
\usepackage{makecell}
\usepackage{array}
\newcolumntype{?}{!{\vrule width 1.8pt}}
\newcommand{\PreserveBackslash}[1]{\let\temp=\\#1\let\\=\temp}
\newcolumntype{C}[1]{>{\PreserveBackslash\centering}p{#1}}
\newcolumntype{R}[1]{>{\PreserveBackslash\raggedleft}p{#1}}
\newcolumntype{L}[1]{>{\PreserveBackslash\raggedright}p{#1}}

\usepackage{xcolor}

\definecolor{patch_1}{rgb}{0.6874,0.1960,0.2278}
\definecolor{patch_2}{rgb}{0.8133,0.4914,0.1612}
\definecolor{patch_3}{rgb}{0.0328,0.5338,0.6474}
\definecolor{patch_4}{rgb}{0.7382,0.3379,0.5850}
\definecolor{patch_5}{rgb}{0.7874,0.6193,0.5439}
\definecolor{patch_6}{rgb}{0.7141,0.5730,0.4868}
\definecolor{patch_7}{rgb}{0.9300,0.1260,0.1533}
\definecolor{patch_8}{rgb}{0.9804,0.9309,0.1332}
\definecolor{patch_9}{rgb}{0.4235,0.7409,0.2779}
\definecolor{patch_10}{rgb}{0.1722,0.6704,0.8863}
\definecolor{patch_11}{rgb}{0.0992,0.1305,0.3152}
\definecolor{patch_12}{rgb}{0.9220,0.0409,0.5469}
\definecolor{patch_13}{rgb}{0.9996,0.9996,0.9996}
\definecolor{patch_14}{rgb}{0.6978,0.7018,0.7174}
\definecolor{patch_15}{rgb}{0.5793,0.5832,0.5988}
\definecolor{patch_16}{rgb}{0.4669,0.4707,0.4865}
\definecolor{patch_17}{rgb}{0.3452,0.3490,0.3569}
\definecolor{patch_18}{rgb}{0.0001,0.0001,0.0001}

\definecolor{patch_13_UWHL}{rgb}{0.15,0.95,0.94}
\definecolor{patch_14_UWHL}{rgb}{0.00,0.83,0.86}
\definecolor{patch_15_UWHL}{rgb}{0.00,0.82,0.86}
\definecolor{patch_16_UWHL}{rgb}{0.00,0.76,0.86}
\definecolor{patch_17_UWHL}{rgb}{0.00,0.58,0.76}
\definecolor{patch_18_UWHL}{rgb}{0.04,0.37,0.51}

\newcommand\RedCell[1]{%
  {\cellcolor{red!15}{#1}}
}

\title{On the limits of perceptual quality measures\\for enhanced underwater images}

\name{Chau Yi Li\sthanks{Email: chauyi.li@qmul.ac.uk}
, Andrea Cavallaro}
\address{Centre for Intelligent Sensing, Queen Mary University of London, UK\\
\\}
\begin{document}
%
\maketitle
\begin{abstract}
The appearance of objects in underwater images is degraded by the selective attenuation of light, which reduces contrast and causes a colour cast. This degradation depends on the water environment, and increases with depth and with the distance of the object from the camera. Despite an increasing volume of works in underwater image enhancement and restoration, the lack of a commonly accepted evaluation measure is hindering the progress as it is difficult to compare methods.  In this paper, we review commonly used colour accuracy measures, such as colour reproduction error and CIEDE2000, and no-reference image quality measures, such as UIQM, UCIQE and CCF, which have not yet been systematically validated. We show that none of the no-reference quality measures satisfactorily rates the quality of enhanced underwater images and discuss their main shortcomings. Images and results are available at \url{https://puiqe.eecs.qmul.ac.uk}.
\end{abstract}

\begin{keywords}
Underwater images, evaluation measures, quality assessment, colour accuracy\end{keywords}

\section{Introduction}

Objects in underwater images are affected by colour cast and reduced contrast due to the selective attenuation of light in water~\cite{Jerlov}. The appearance degradation depends on the water type: in  oceanic  waters red light is most attenuated and objects are under a blue colour cast; in coastal  waters blue light is most attenuated and objects are under a green or yellow colour cast~\cite{Jerlov}.
Despite an increasing interest in underwater image enhancement and restoration~\cite{WaterNET_TIP2019,Anwar_Li_2020_Survey}, no standard evaluation measure or protocol exists for assessing the quality of the enhanced underwater images. Perceptual quality can be assessed with a panel of human observers performing subjective tests. Unlike traditional image quality assessment~\cite{ITURec500-14}, subjective tests for underwater images have not been standardised yet. In studies reported in the literature, subjects are given a specific task, such as choosing their {\em preferred} images~\cite{BglightEstimationICIP2018_Submitted}, ranking multiple images~\cite{ Anwar2018DeepUI_UWCNN}, and assessing a particular feature of the image. Examples include choosing the image that {\em looks more like it is taken under a white illuminant}~\cite{Li_2020_Cast-GAN},  scoring images based on a certain criteria, such as {\em how realistic they are}~\cite{Protasiuk2019_localcolourmapping}, or indicating whether the {\em colour patches on a colour checker can be distinguished}~\cite{Zhang2017_multisclae_retinex}. Conducting subjective tests is time-consuming as they need a diverse pool of participants who shall be appropriately trained for the task~\cite{ITURec500-14}.  

As an alternative to subjective tests, objective measures can be used to quantify perceptual quality. Colour accuracy measures can be used to quantify the perceived colour difference between the enhanced image and a reference image, hence quantifying the ability of an algorithm to remove colour degradations. These measures need a reference image or object, such as a colour checker with known colours under a reference illuminant~\cite{Ancuti2018, BermanArvix_UWHL_dataset}. Two publicly available datasets, namely SQUID~\cite{BermanArvix_UWHL_dataset} and Sea-thru~\cite{seathru2019CVPR}, contain underwater images with colour checkers. 
Image quality assessment (IQA) measures for underwater images quantify colour and contrast degradations. These measures combine image attributes to mimic human preferences in the enhanced images. It has been widely reported that underwater-specific IQA measures~\cite{UIQM_Panetta2016, UCIQE, WANG2018_NR_CFF} do not reflect subjective judgement~\cite{BermanArvix_UWHL_dataset, WaterNET_TIP2019, BglightEstimationICIP2018_Submitted, Ucolor}. 

In this paper, we review perceptual quality measures for underwater images and discuss their limitations. To the best of our knowledge, this is the first systematic analysis of underwater evaluation measures. We show that the colour reproduction accuracy measure~\cite{BermanArvix_UWHL_dataset} is inadequate to quantify colour difference and that current IQA measures~\cite{BermanArvix_UWHL_dataset, WaterNET_TIP2019,Ucolor} fail to account for visible artefacts, such as red overcompensation. Moreover, we show that numerical changes in the measure do not relate to perceptual quality differences.  We complement this work with a online platform\footnote{\url{https://puiqe.eecs.qmul.ac.uk/}} with underwater enhancement methods to facilitate comparisons.

\section{Quality Assessment}
Objective measures commonly used to assess the quality of enhanced underwater images include colour accuracy and IQA measures.  These measures can be generic (i.e.~designed for images taken in air) or specifically designed for underwater settings. A measure may use an ideal, reference image or  object (full-reference measure) or operate without a reference image or object (no-reference measure).

Using a reference object, {\em colour accuracy} can be measured as a function of the difference between the appearance of a colour checker in the enhanced image and that of the checker under a reference illuminant~\cite{sharma2005ciede2000}. The colour reproduction error~\cite{BMVCAngularError} was originally proposed for illuminant estimation algorithms and measures the angle between vectors representing the true and estimated white (i.e.~the white surface under the unknown light mapped to reference light using an estimated illuminant in the RGB colour space). The colour reproduction error has  been used to measure the colour accuracy of the achromatic patches of the colour checker in underwater images~\cite{BermanArvix_UWHL_dataset}. Colour accuracy has also been measured as the distances in the RGB~\cite{Li2018WaterGANUG, BglightEstimationICIP2018_Submitted} and CIELab colour spaces~\cite{Ancuti2018, Ucolor}. A Euclidean distance in RGB is not representative of the colour difference perceived by human, as the RGB space is not perceptually uniform~(i.e.~a certain numerical difference in different parts of the space is similarly perceived as the change in colour).  CIEDE2000~($\Delta E_{00}$) is a more appropriate distance measure of two colour samples in the CIELab colour space and is defined as~\cite{sharma2005ciede2000}:
\begin{equation}
     \Delta  E_{00} \!=\! \sqrt{\left(\frac{\Delta L}{k_L S_L}\right)^{\!\!2}\!+\!\left(\frac{\Delta C}{k_C S_C}\right)^{\!\!2}\!+\!\left(\frac{\Delta H}{k_H S_H}\right)^{\!\!2}\!+\!\Delta R},
\end{equation}
where~\mbox{$\Delta L, \Delta C, \Delta H$} are the differences in lightness, chroma and hue of the colour samples,~\mbox{$S_L,S_C,S_H$} are weighing functions that depends on the location of the samples in the colour space, and~$k_L, k_C, k_H$ are weighing factors that vary according to the viewing conditions. The term~$\Delta R$ aims to improve the chromatic differences performance for the blue region. \mbox{$\Delta  E_{00} \in [0, 100]$}, and~\mbox{$\Delta  E_{00} \leq 1$} corresponds to a visually imperceptible  difference.

 {\em Generic} IQA measures quantify the reconstruction error or desirable properties of an image. The reconstruction error can be measured by mean square error (MSE), peak signal-to-noise ratio (PSNR) or structural similarity index measure (SSIM)~\cite{wang2004ssim}. Most methods use the original (degraded)  image as the reference~\cite{Zhang2017_multisclae_retinex} or an  enhanced image that was selected by the panel of a subjective test~\cite{WaterNET_TIP2019,Ucolor}. 
 Desirable image properties include sharpness, contrast and naturalness. Sharpness and contrast can be measured with a visible-edge count and entropy~\cite{ICIP2017UWCNN}. These measures may be affected by noise introduced by the enhancement method itself. Naturalness can be measured with BRISQUE~\cite{BRISQUE}, which quantifies distortions such as noise and blur. 

{\em Underwater-specific} measures quantify the perceived image quality via attributes related to degradations in water. The attributes (e.g.~colour,  contrast and visibility) may be combined linearly with weights derived from subjective tests on underwater images. Two full-reference measures, namely~$Q_u$~\cite{Lu2016_descattering_quality_ICIP} and~PKU-EAQA~\cite{PKU}, and three no-reference measures, namely   Underwater Color Image Quality Evaluation~(UCIQE)~\cite{UCIQE},  Underwater Image Quality Measure~(UIQM)~\cite{UIQM_Panetta2016} and Colorfulness, Contrast and Fog density index~(CCF)~\cite{WANG2018_NR_CFF}, are commonly used as the evaluation measures.
Table~\ref{Table: evaluation_measures_obj} summaries the measures and their subjective test details. 
For the full-reference measures,~$Q_u$ measures the colour difference in the CIELab colour space~\cite{sharma2005ciede2000} and SSIM~\cite{wang2004ssim} between the reference image and the enhanced image. The attributes were heuristically combined with equal weights without validation by a subjective test. PKU-EAQA~\cite{PKU} compares the relative quality between two images, using attributes of colour (colour motion~\cite{Stricker1995}) and contrast (GIST~\cite{Oliva2001} descriptor). 
Although PKU-EAQA allows the comparison between several enhanced methods through successive comparison, it does not state whether a method improves the image quality marginally or by a large extent than an other method.

\begin{table}[t!]
\caption{Underwater-specific IQA measures and their associated subjective tests. The measures are grouped into full-reference (Yes) or no-reference (No), based on whether they need a reference image (Ref.). The measure can be applied (App.) to score an image or to rank the relative quality between two images. Attributes of the measures are colour (Col.) and/or visibility (Vis.). We list the number of images (Img.), the number of participants (Part.), and the task the participants were asked to perform in the subjective tests. }
\vspace{5pt}
\scriptsize
\centering
\setlength\tabcolsep{1pt}
\begin{tabular}{llllccrrL{3.5cm}}
\multicolumn{1}{c}{\textbf{Ref.}} & 
\multicolumn{2}{c}{\textbf{Measure}} &\multicolumn{1}{c}{\textbf{App.}} &\multicolumn{2}{c}{\textbf{Attributes}}  &\multicolumn{3}{c}{\textbf{Subjective Test}}\\
\multicolumn{2}{c}{\textbf{}} &\multicolumn{2}{c}{} &\multicolumn{1}{c}{Col.}&\multicolumn{1}{c}{{Vis.}} &  {{Img.}}  & {{Part.}} & \multicolumn{1}{c}{Task} \\
\hline
\multirow{2}{*}{Yes} 
&\cite{Lu2016_descattering_quality_ICIP} & $Q_u$  & score & \checkmark &\checkmark  &\multicolumn{3}{c}{\cellcolor{black!40}N/A}\\
&\cite{PKU}  & PKU-EAQA  &  rank &  \checkmark &\checkmark &  400  & 30 & determine `{\em which of the two is better}'\\
\hline
\multirow{3}{*}{No} 
&\cite{UCIQE} & UCIQE & score &  \checkmark &   & 44 & 12 & score from `Very annoying' to  `Imperceptible'\\

&\cite{UIQM_Panetta2016} & UIQM & score &  \checkmark &\checkmark  & 126  & 10 & not disclosed \\
&\cite{WANG2018_NR_CFF} & CCF & score &  \checkmark
&\checkmark   & 87 &  {{20}} & score from `Bad' to `Excellent'\\
\end{tabular}
\label{Table: evaluation_measures_obj}
\vspace{-15pt}
\end{table}

No-reference IQA measures are generally used to compare the results of different enhancement methods, because of the lack of appropriate reference images. UCIQE~\cite{UCIQE} uses attributes related to colour  degradations in the CIELab colour space~\cite{CIELab}, namely standard deviation in chroma, average in saturation and difference between extreme values (i.e.~the top and bottom 1\% in luminance). The chroma and saturation statistics only focus on individual pixel intensities but not on the spatial distribution of intensities, whereas the difference between extreme pixels can be biased towards distortions such as over-exposure in one region. UIQM~\cite{UIQM_Panetta2016} accounts for the colour degradation considering colourfulness (using the opponent colour theory) and contrast (edge sharpness and contrasts)  as attributes. CCF~\cite{WANG2018_NR_CFF} uses a colourfulness index and measures blur and (lack of) visibility with a contrast measure and a foggy index. The images used in the subjective test were taken in a water tank with different concentrations of aluminium hydroxide, which gives a white solution, to mimic different levels of turbidity.   
 
\begin{figure*}[t!]
\setlength\tabcolsep{0.5pt}
\centering
\scriptsize
\begin{tabular}{ccccccccccccc}
\includegraphics[width=0.098\textwidth]{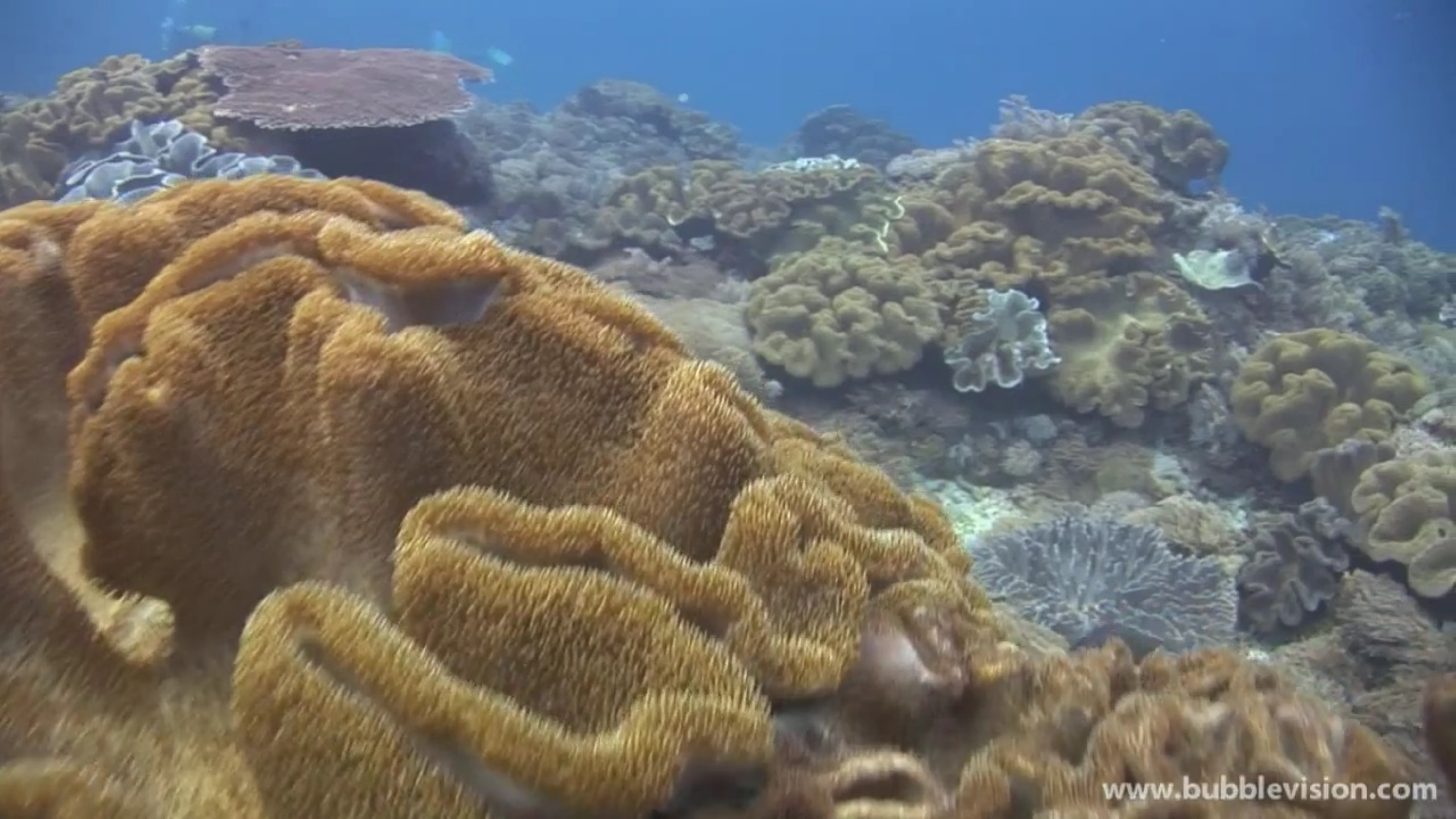} & \includegraphics[width=0.098\textwidth]{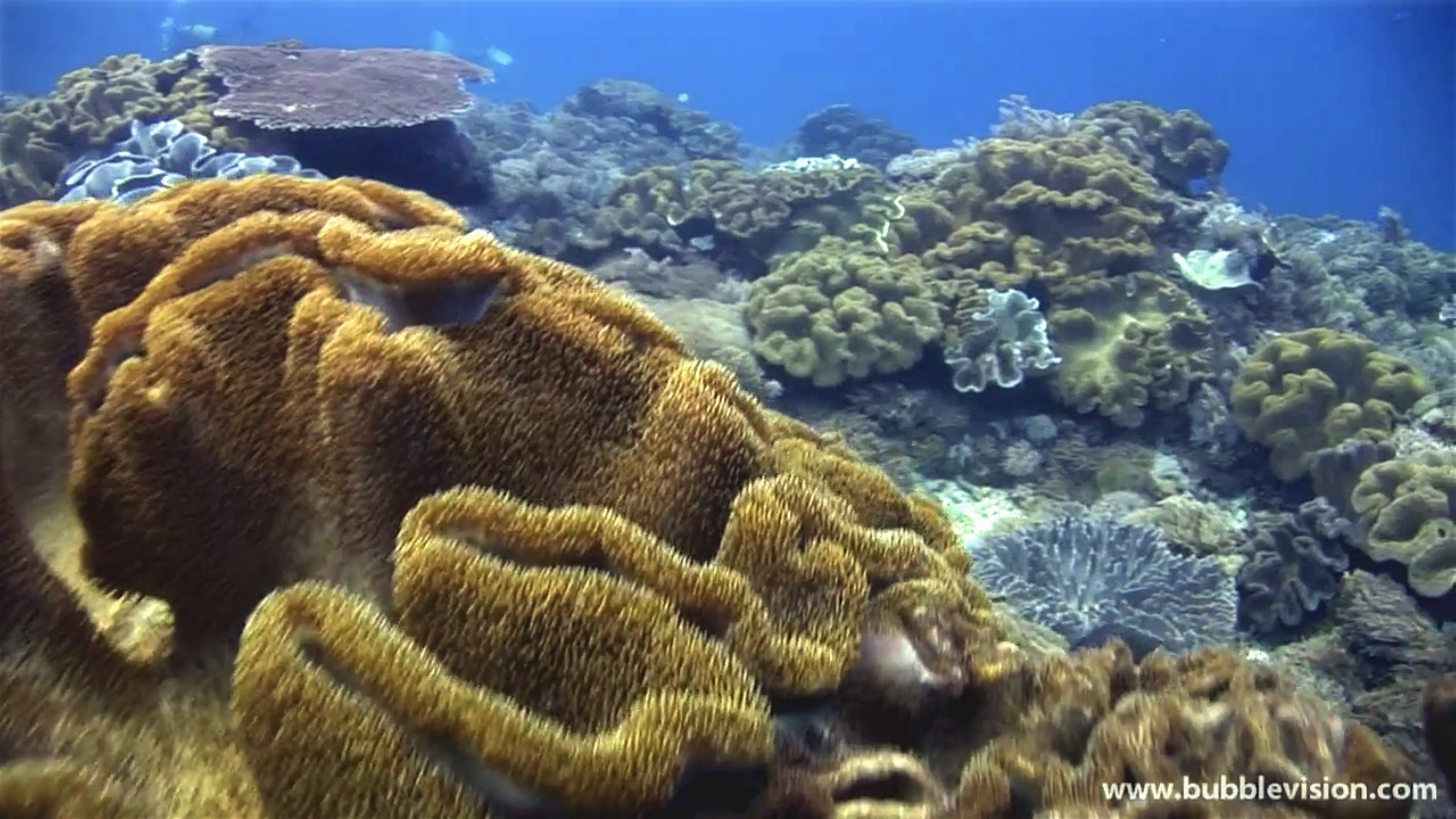} & \includegraphics[width=0.098\textwidth]{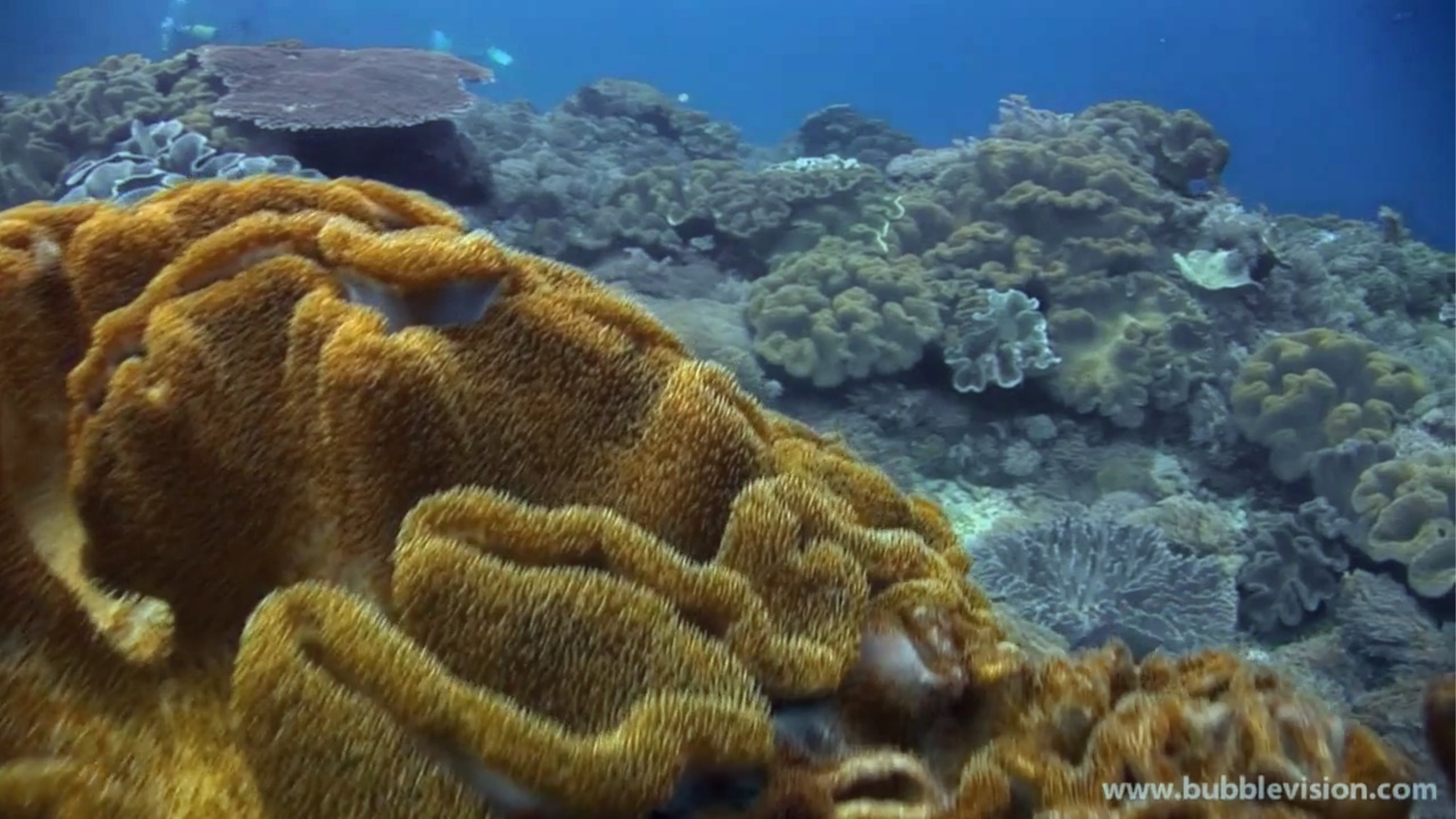} & \includegraphics[width=0.098\textwidth]{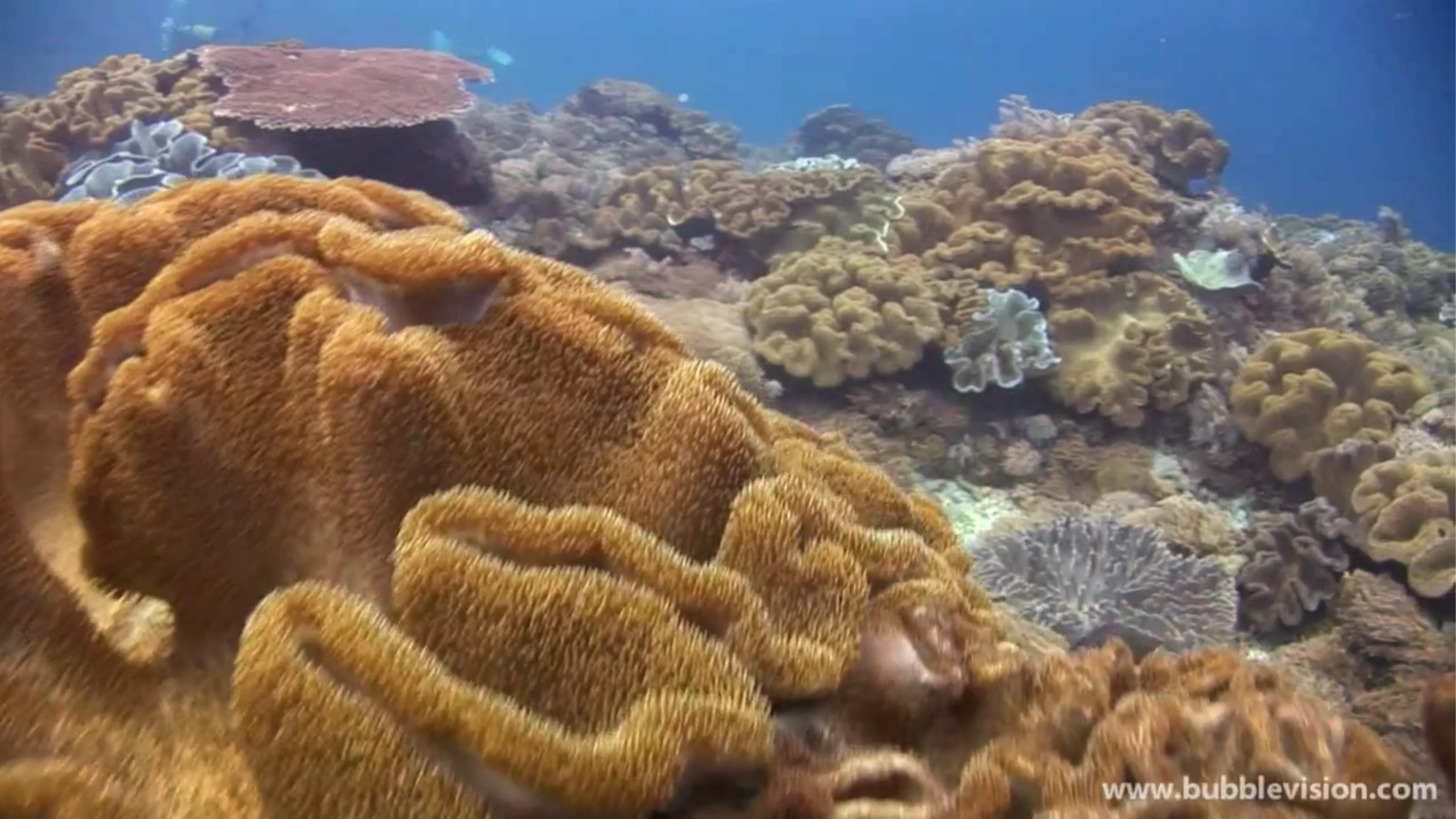} & \includegraphics[width=0.098\textwidth]{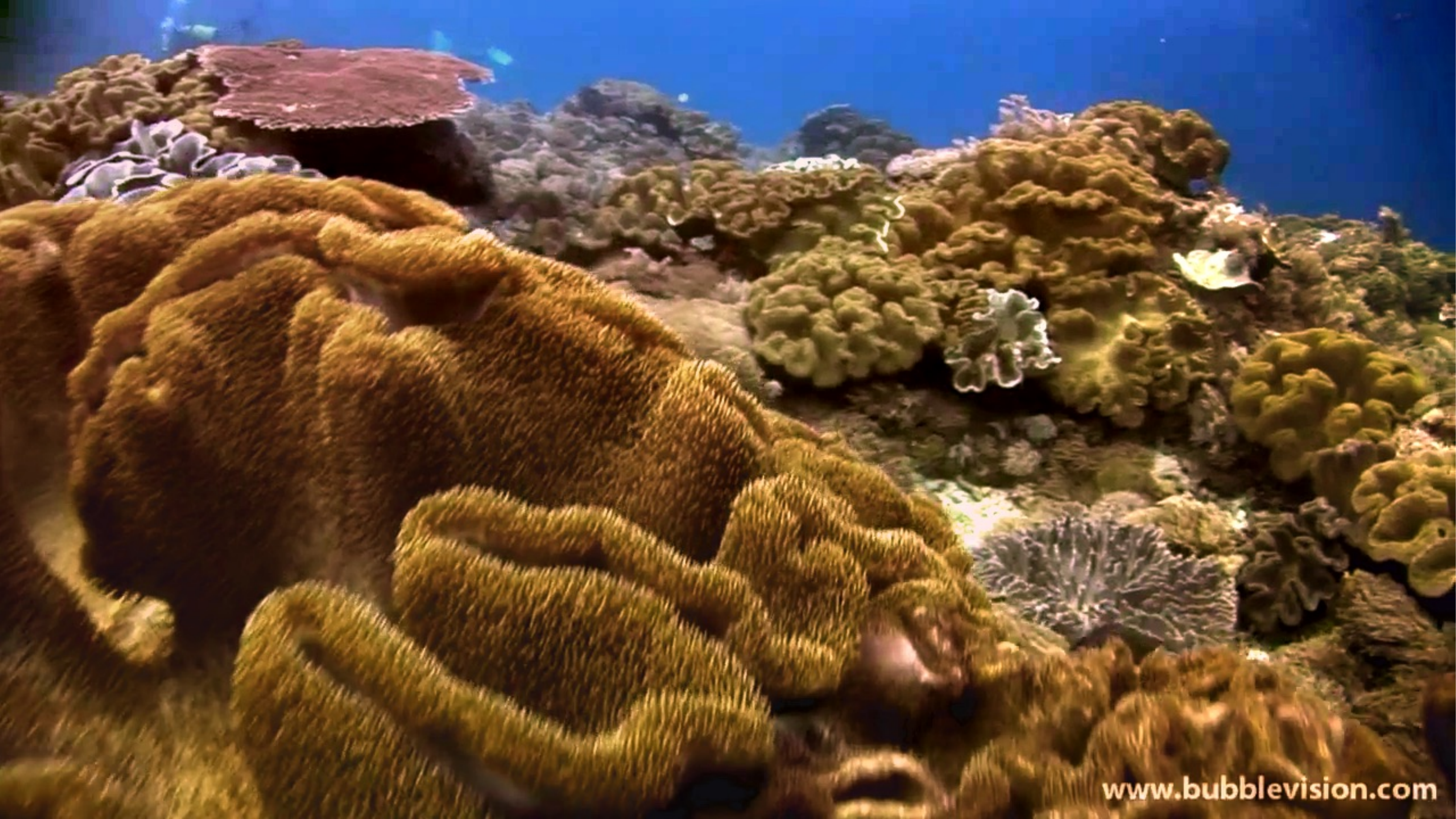} & \includegraphics[width=0.098\textwidth]{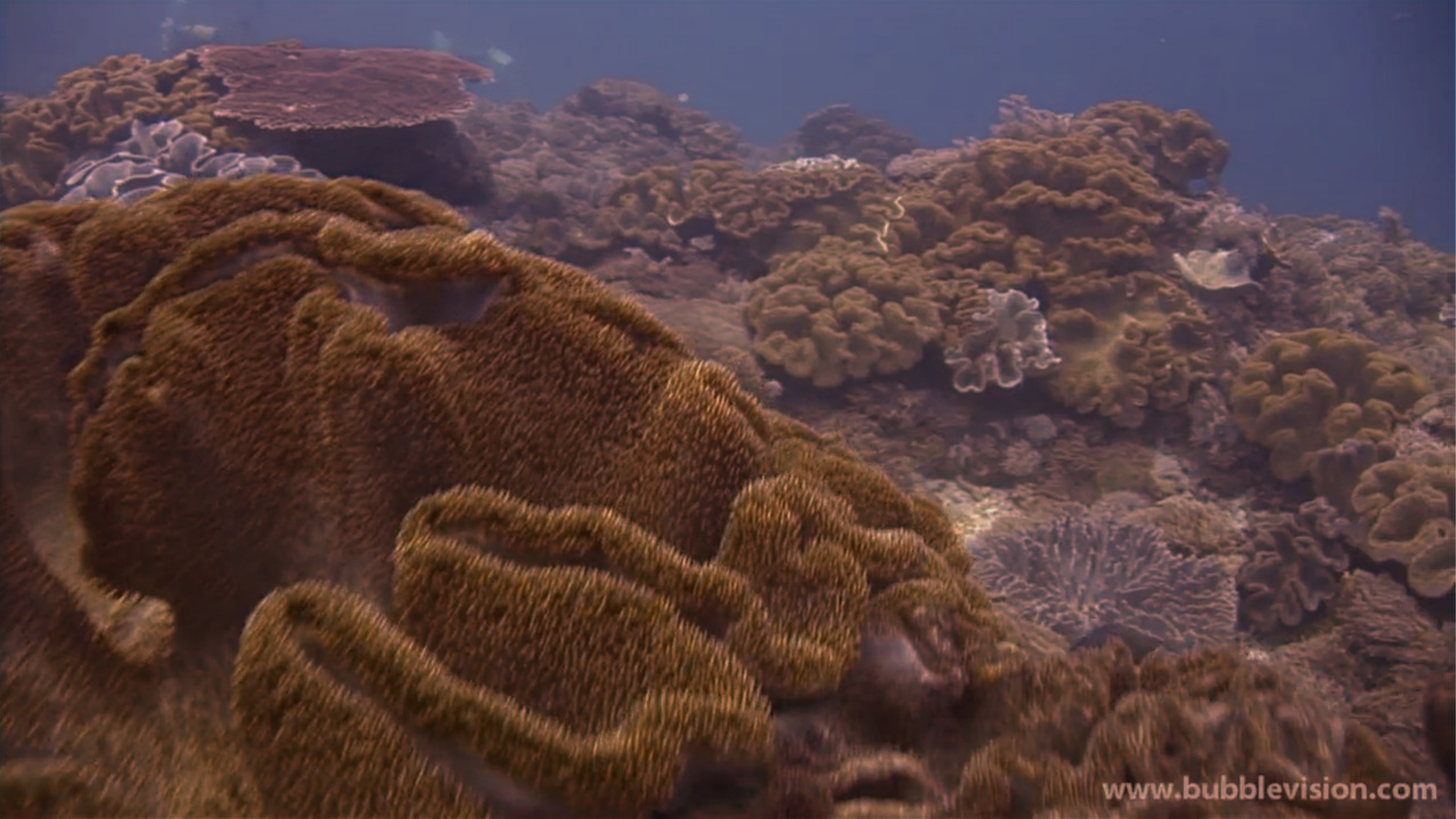}  & \includegraphics[width=0.098\textwidth]{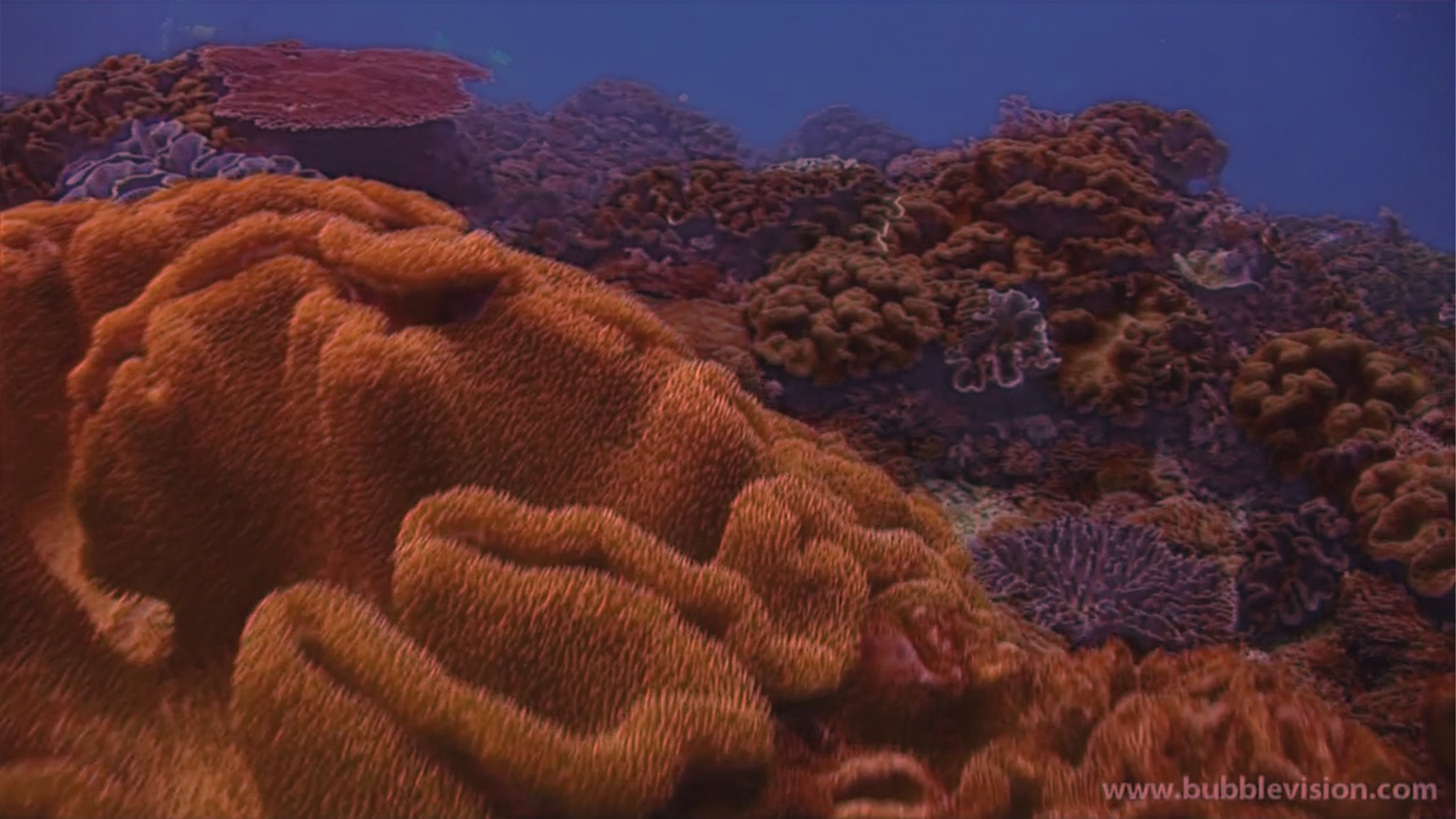} & \includegraphics[width=0.098\textwidth]{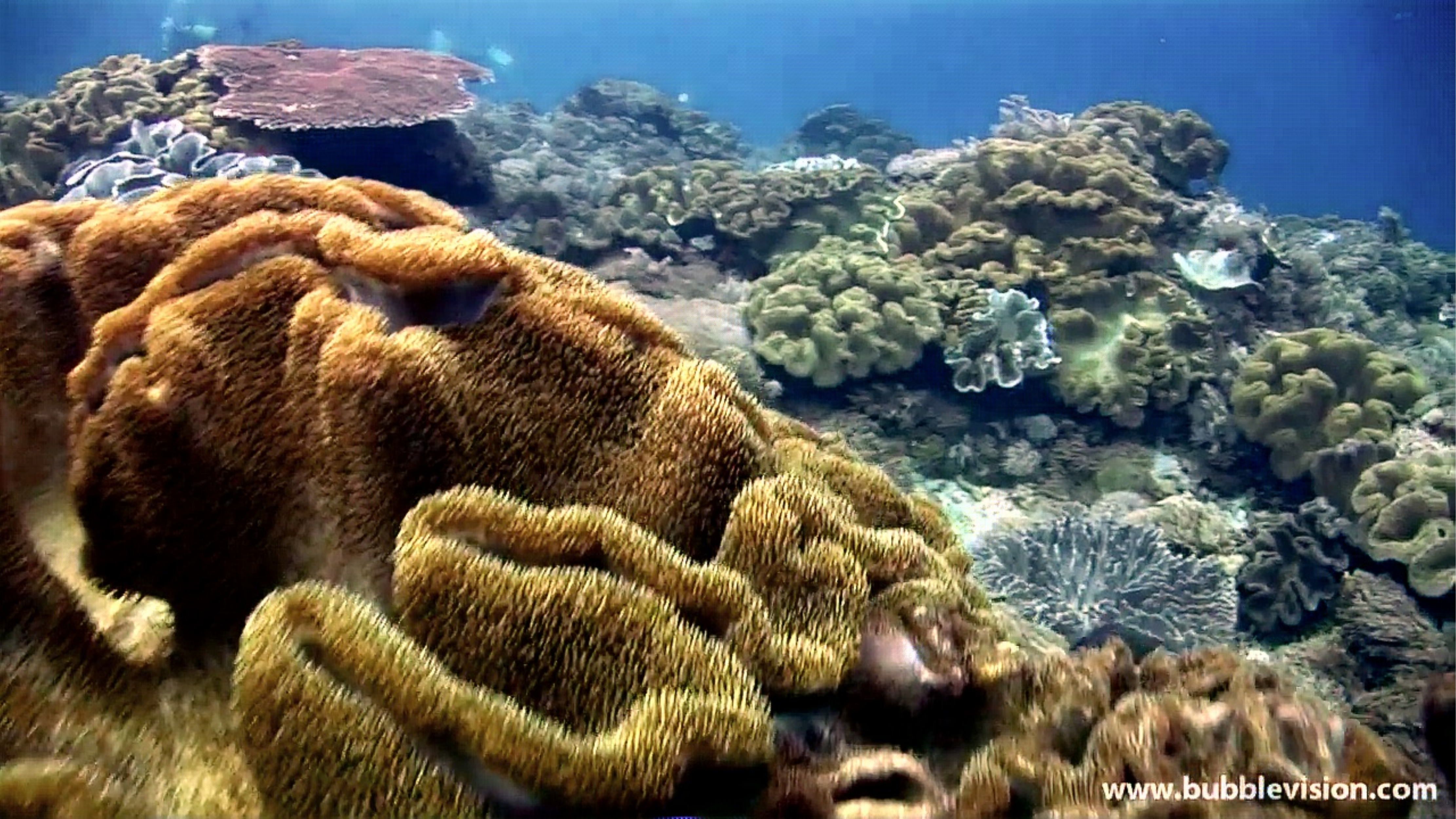} &\includegraphics[width=0.098\textwidth]{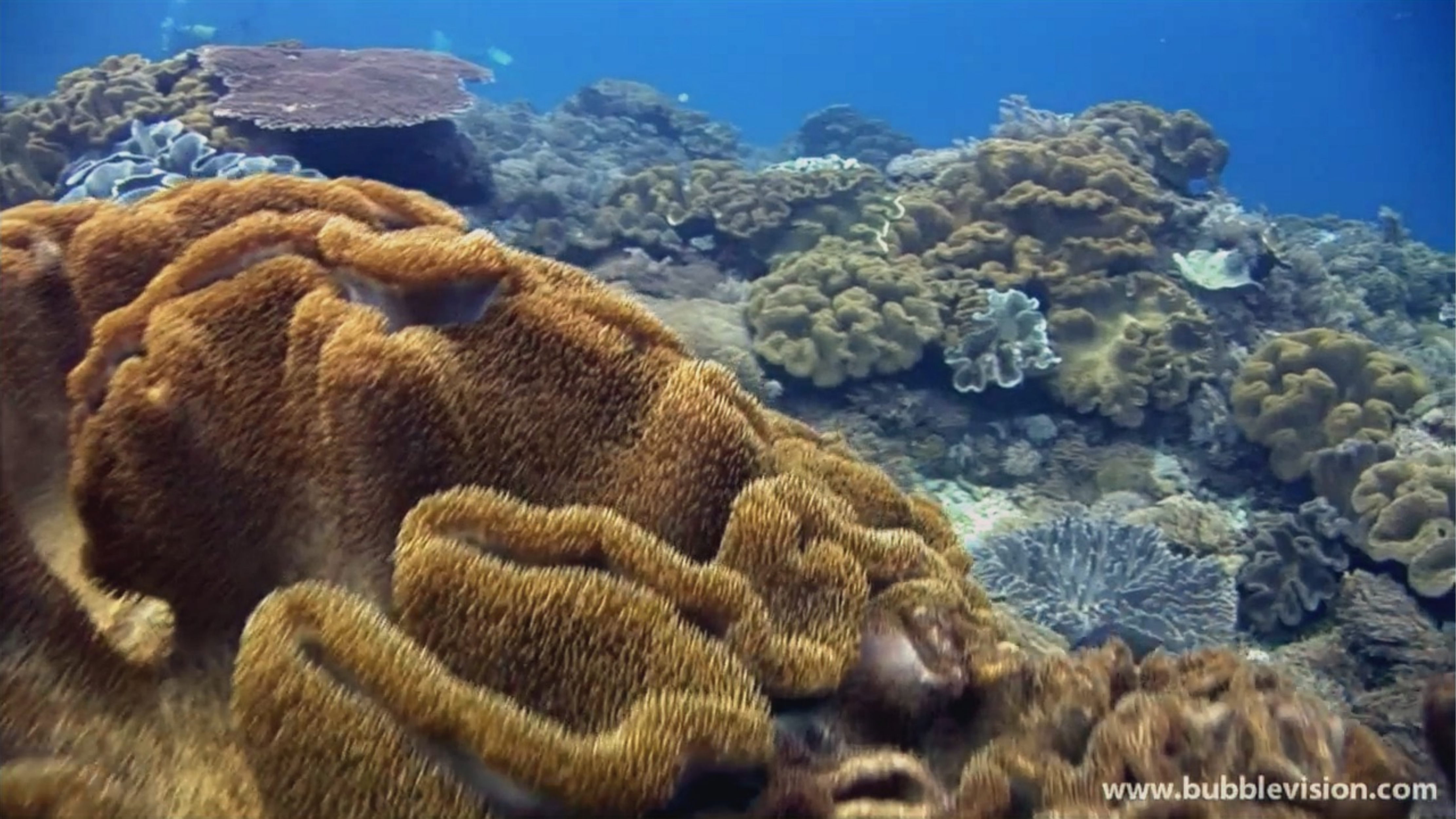} & \includegraphics[width=0.098\textwidth]{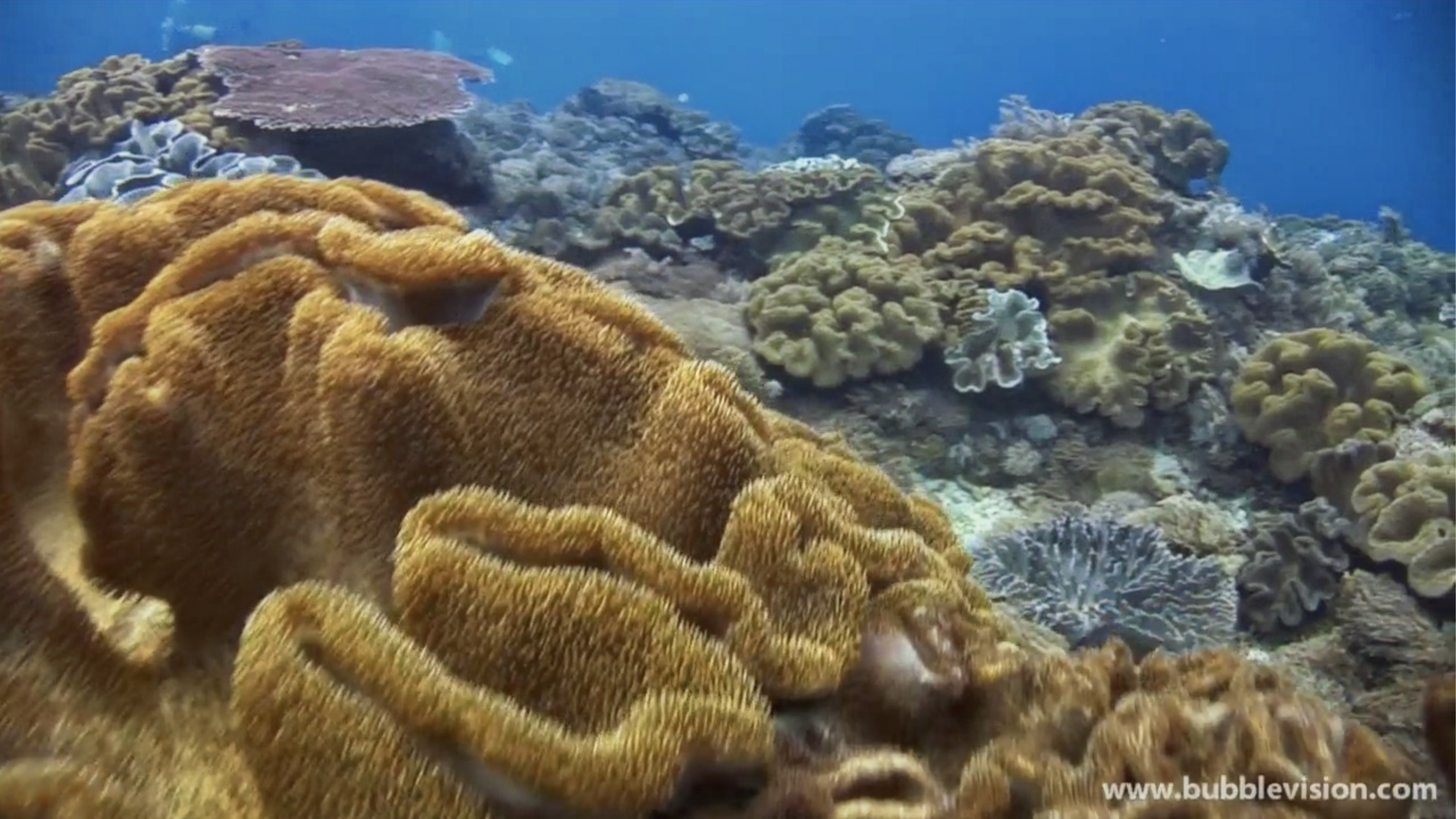}  \\
\includegraphics[width=0.098\textwidth]{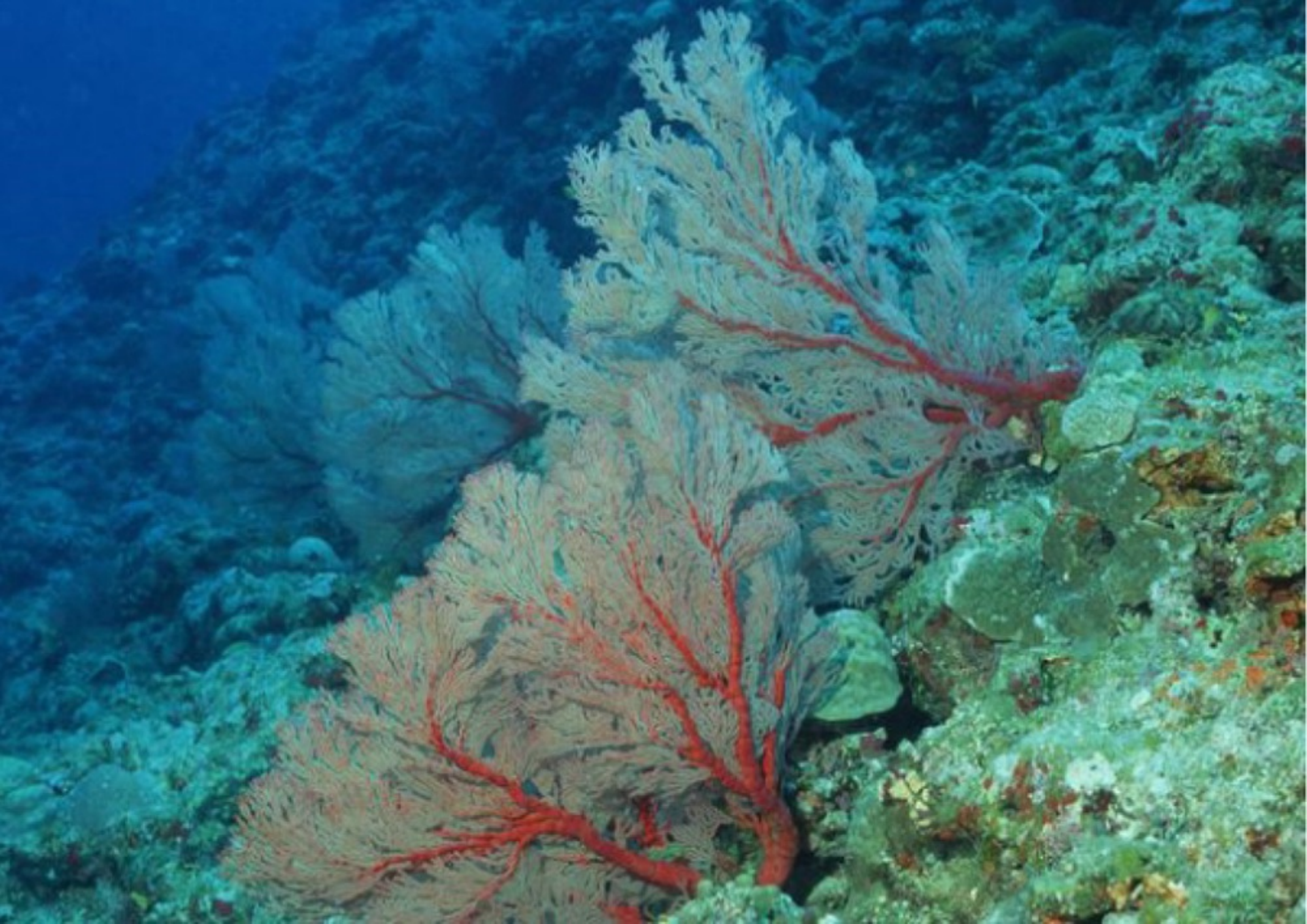} & \includegraphics[width=0.098\textwidth]{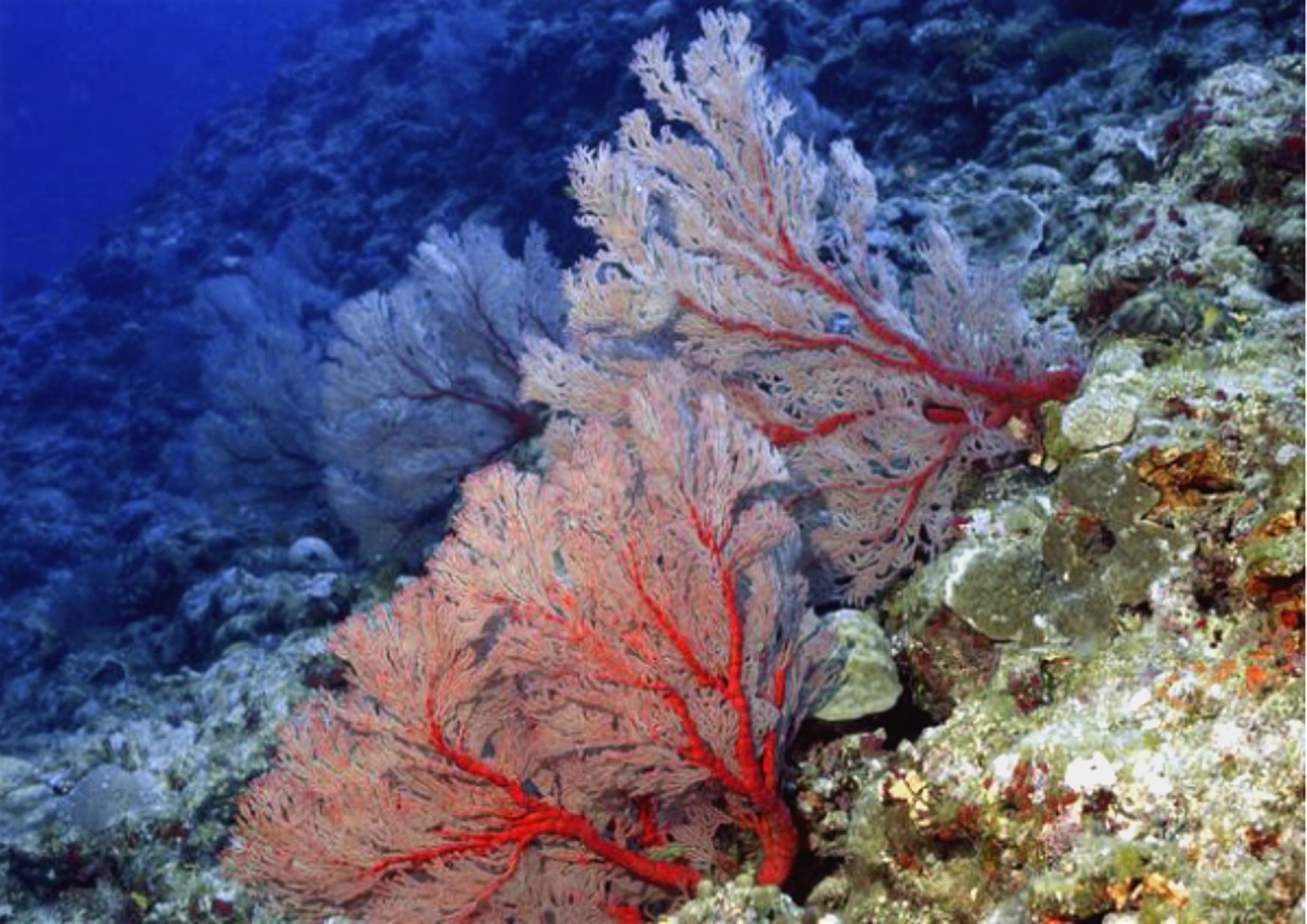} & \includegraphics[width=0.098\textwidth]{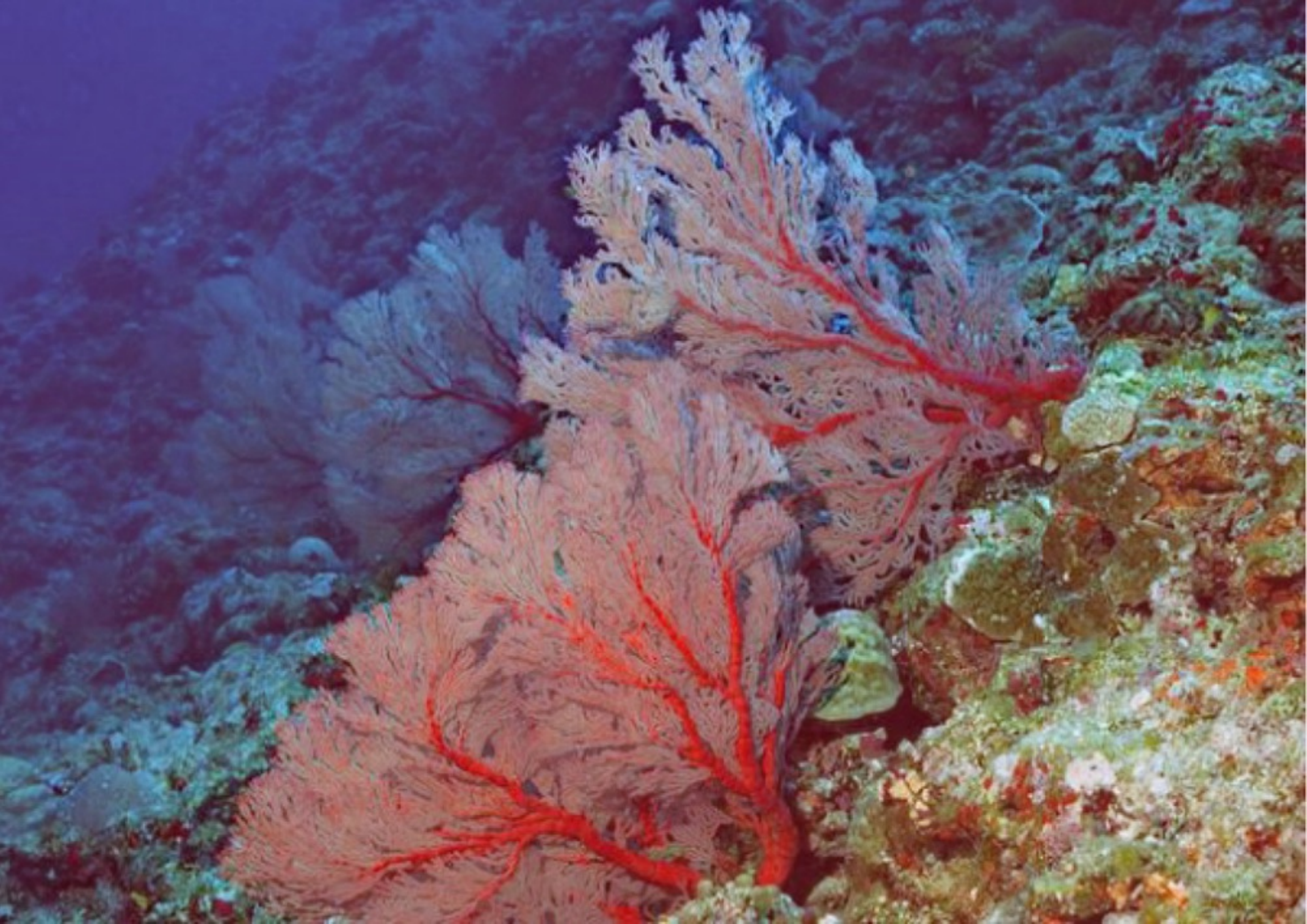} & \includegraphics[width=0.098\textwidth]{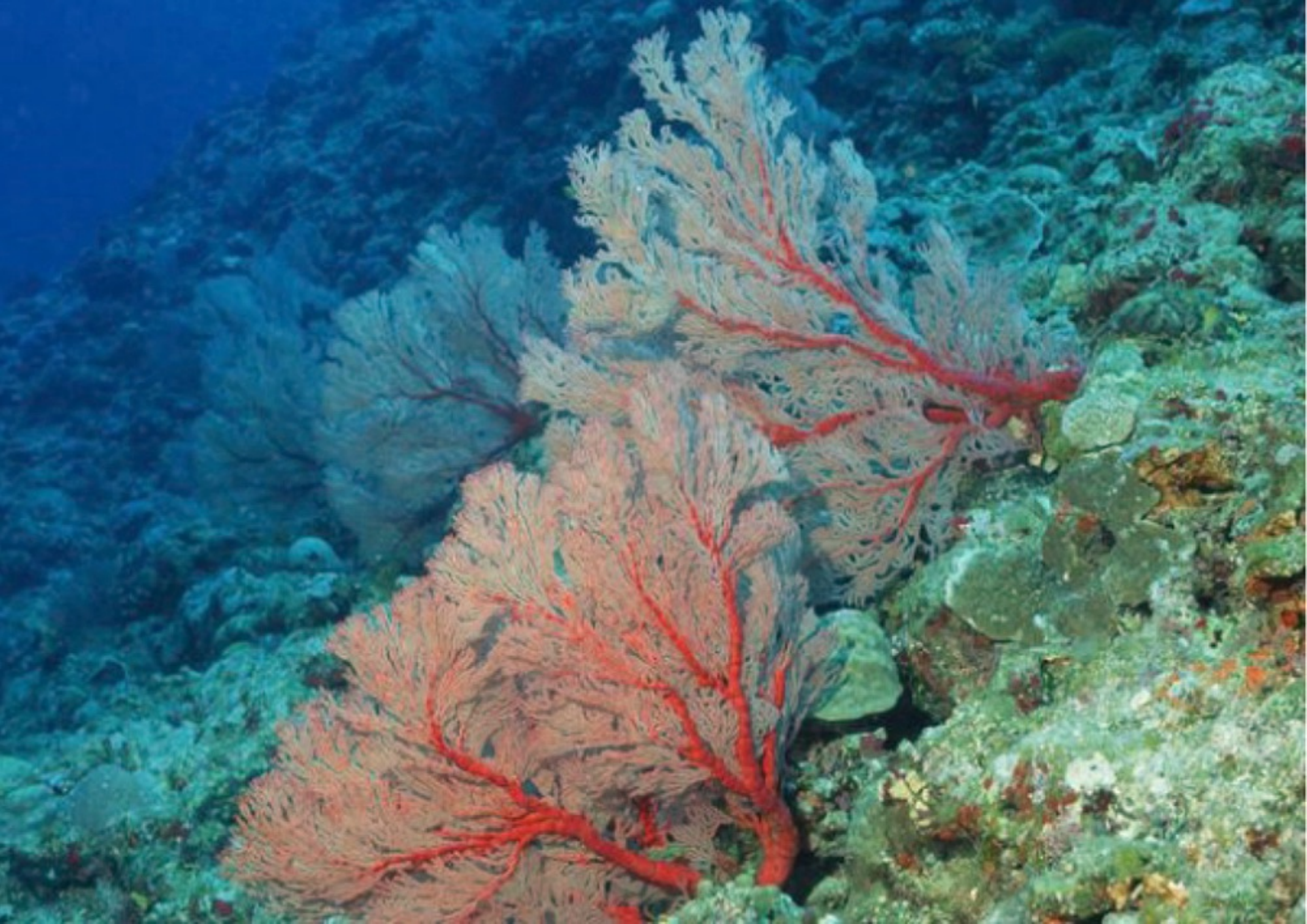} & \includegraphics[width=0.098\textwidth]{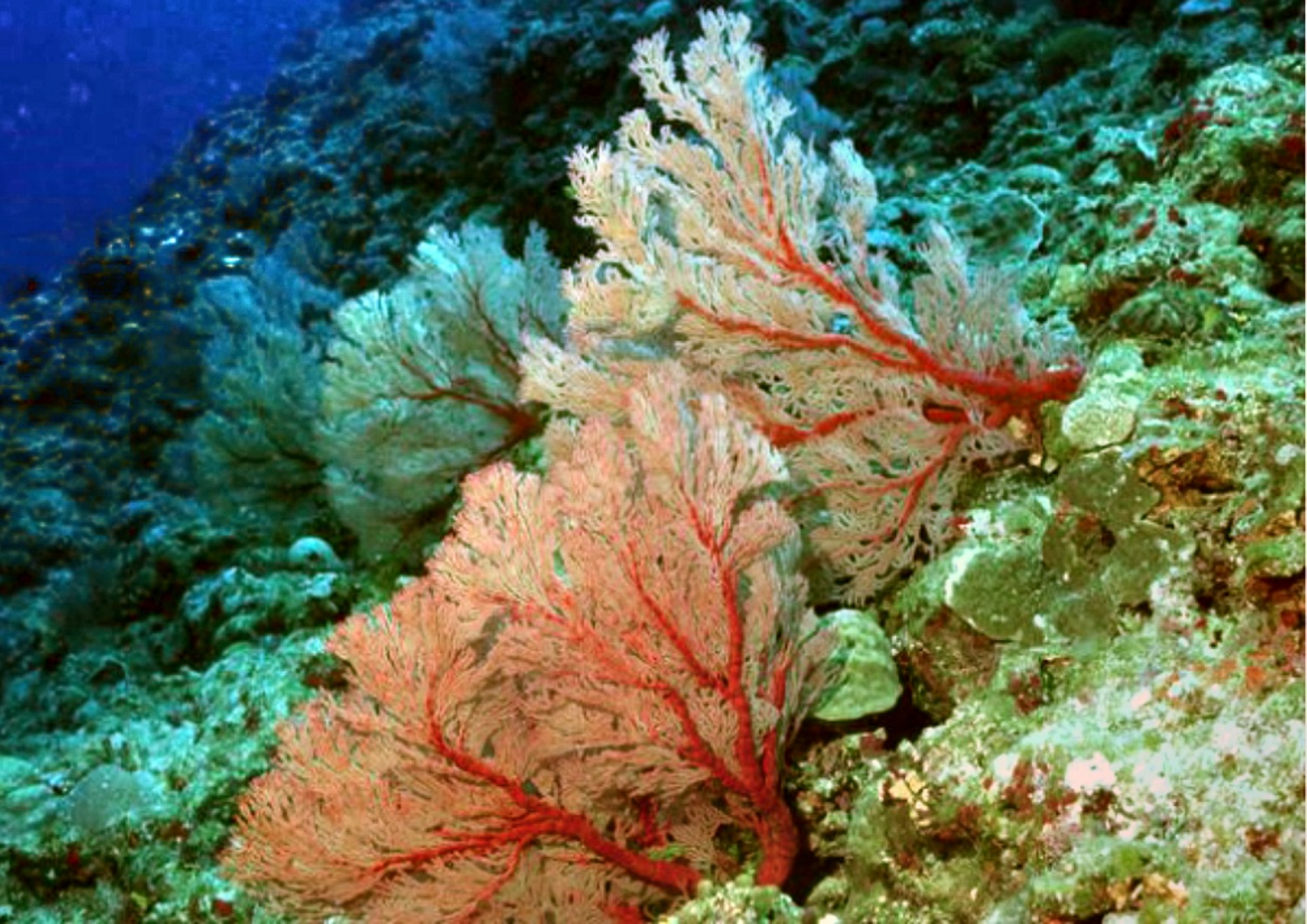} & \includegraphics[width=0.098\textwidth]{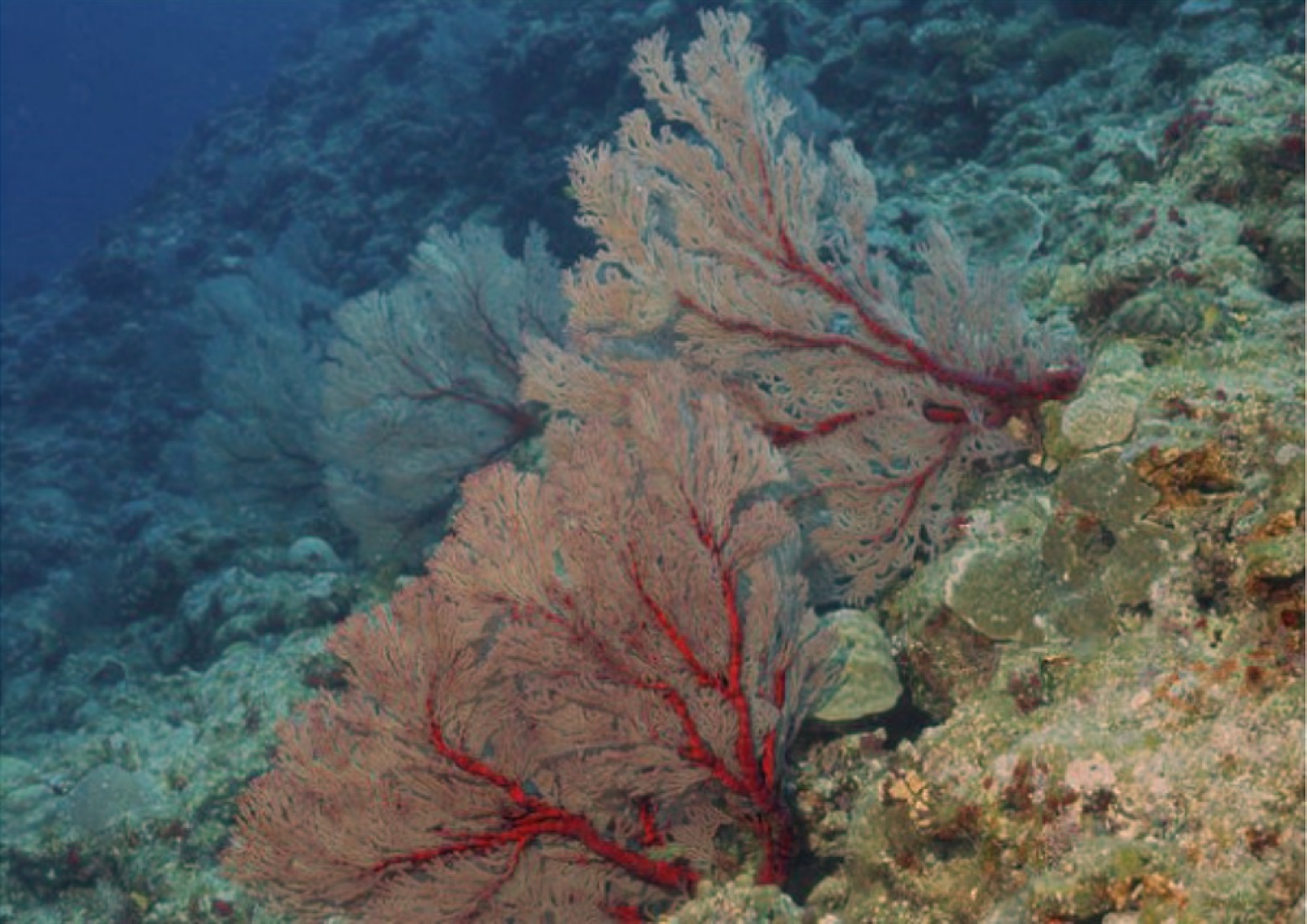}  & \includegraphics[width=0.098\textwidth]{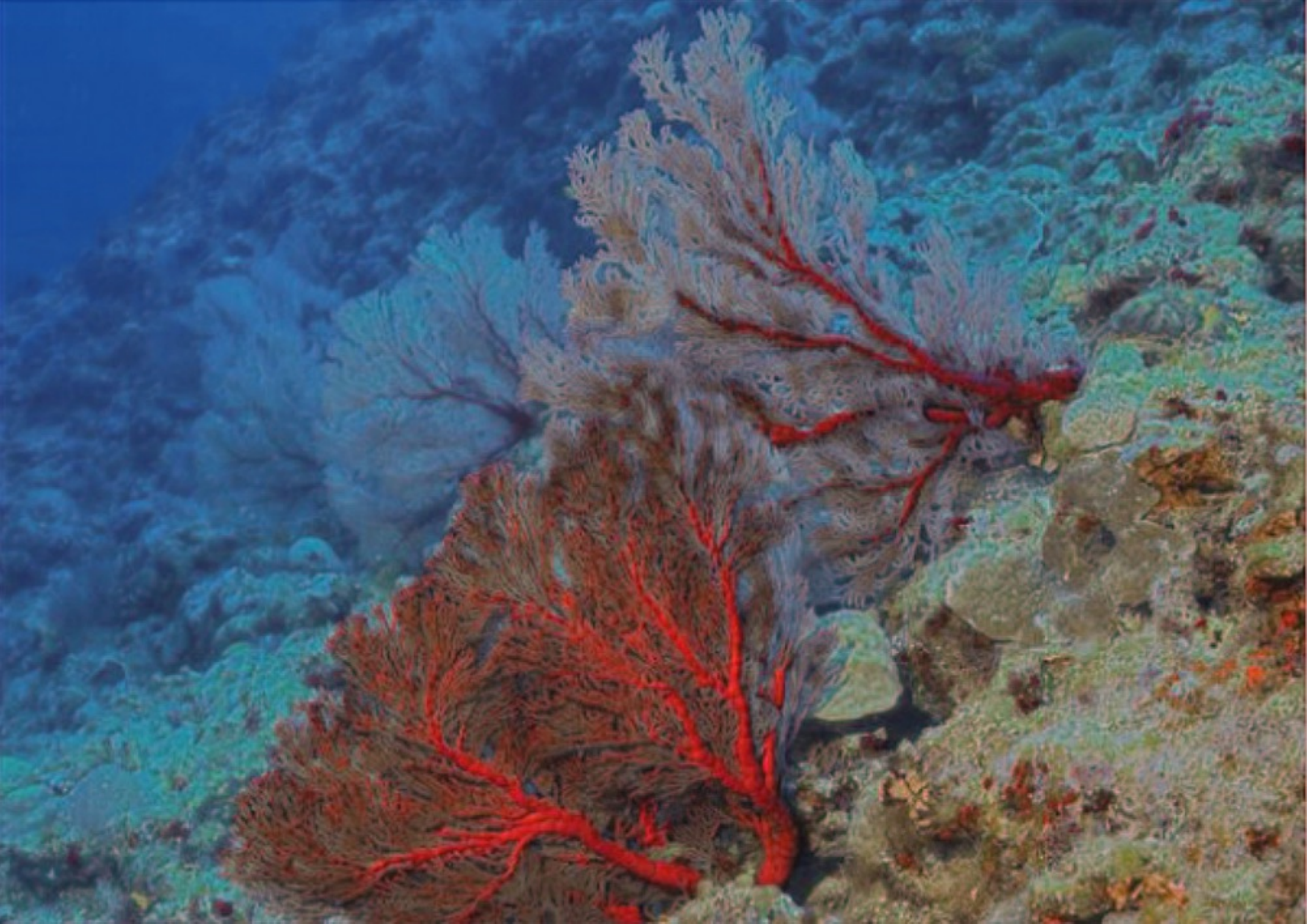} &\includegraphics[width=0.098\textwidth]{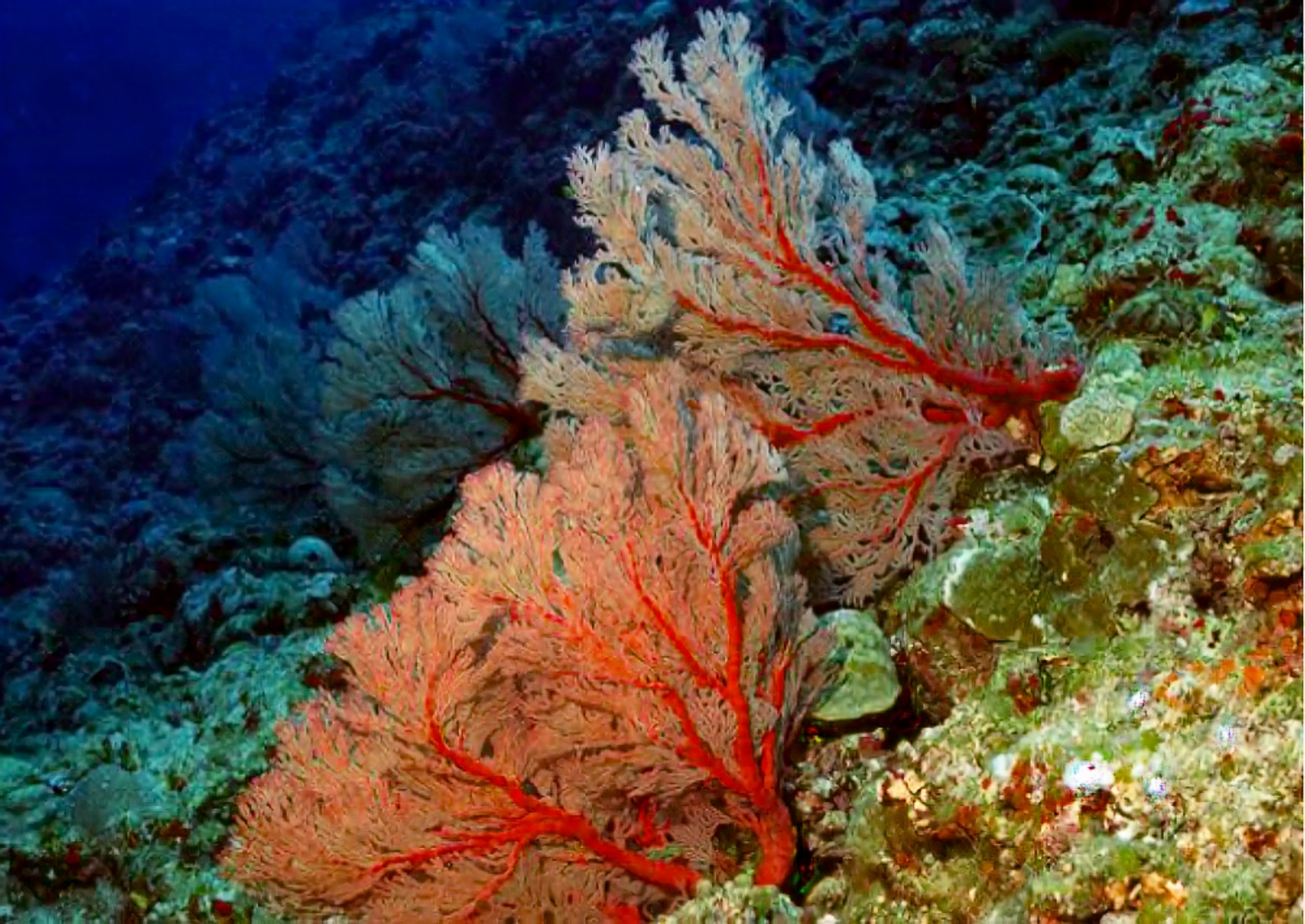} & \includegraphics[width=0.098\textwidth]{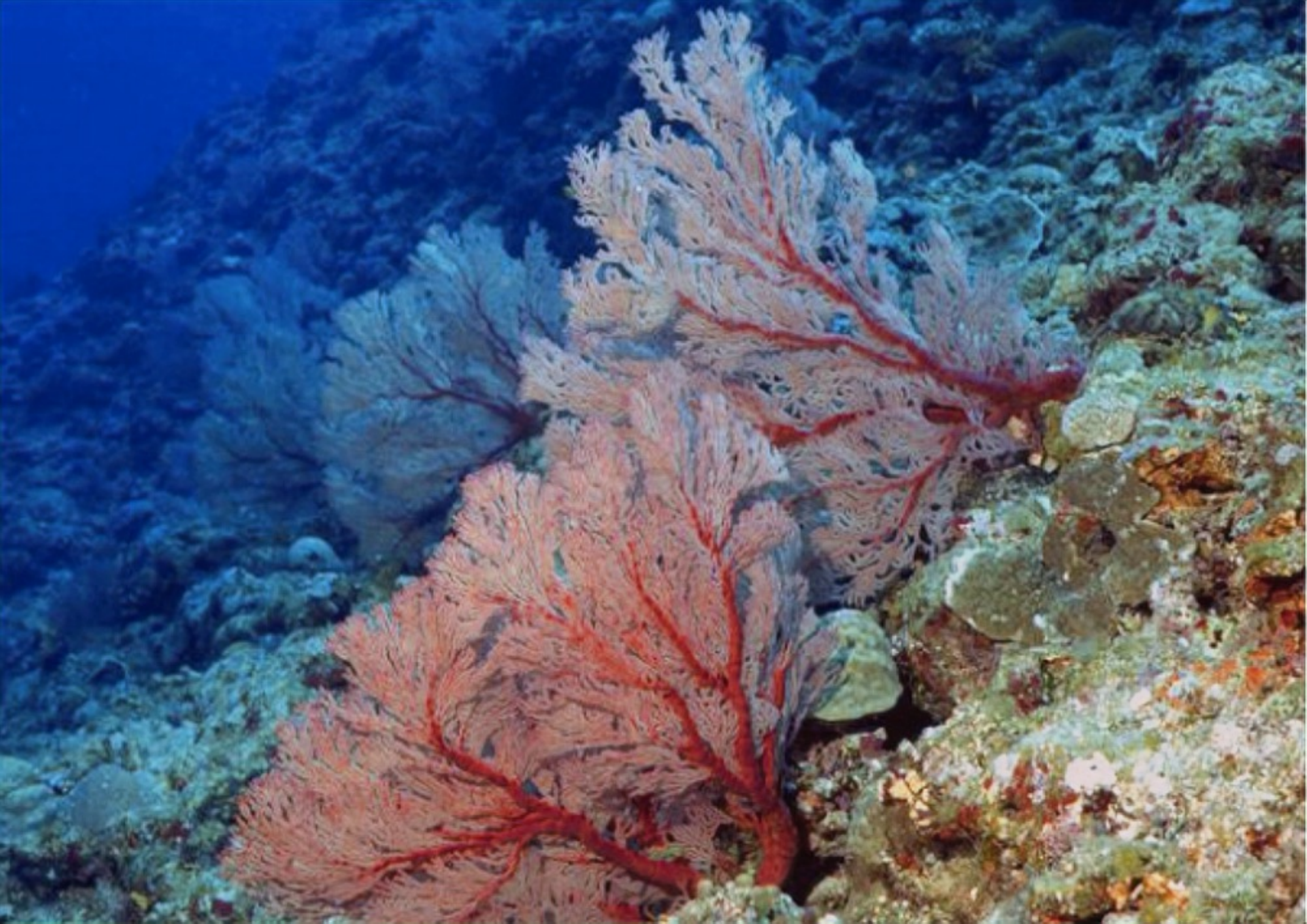} &  \includegraphics[width=0.098\textwidth]{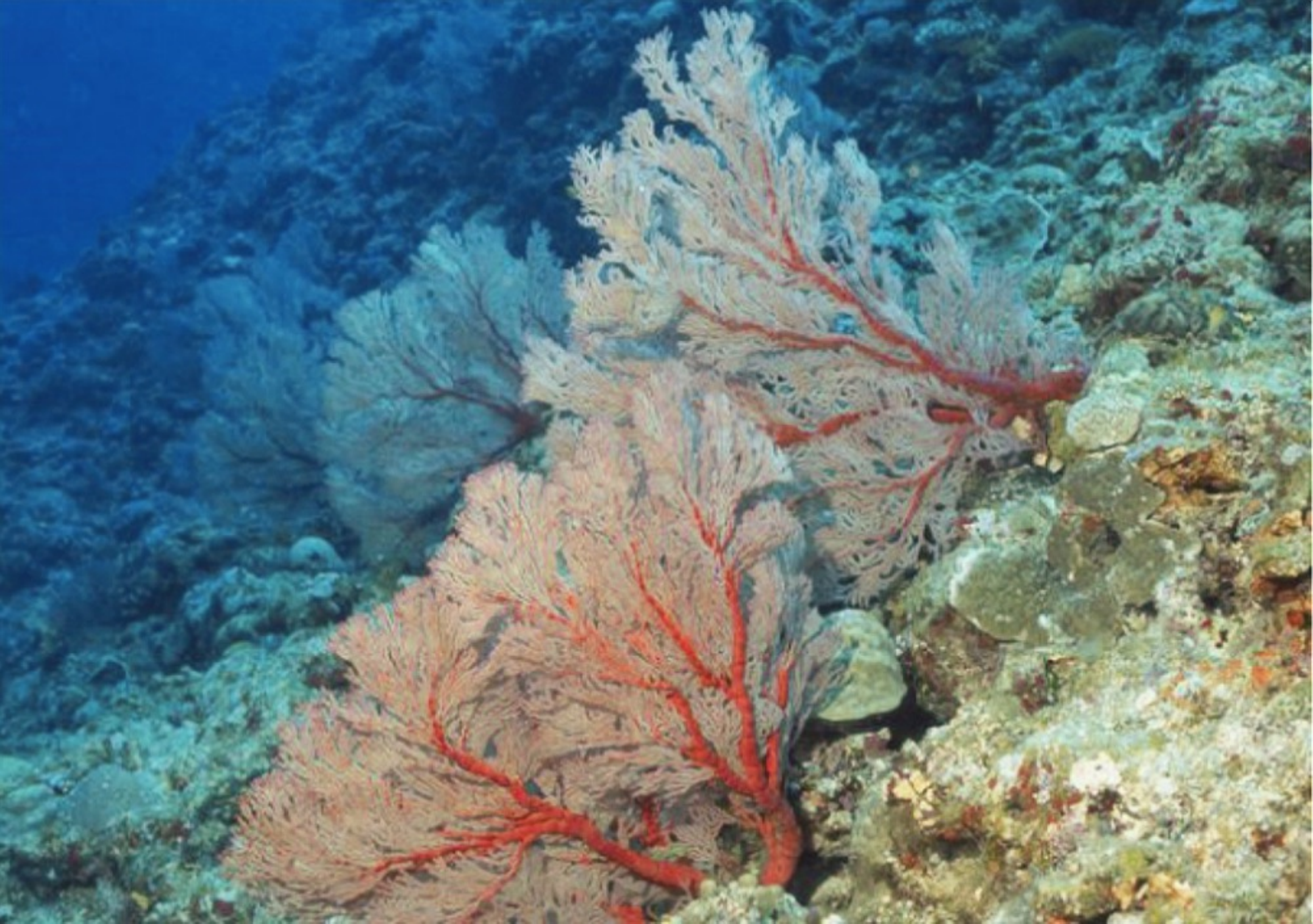}\\
\includegraphics[width=0.098\textwidth]{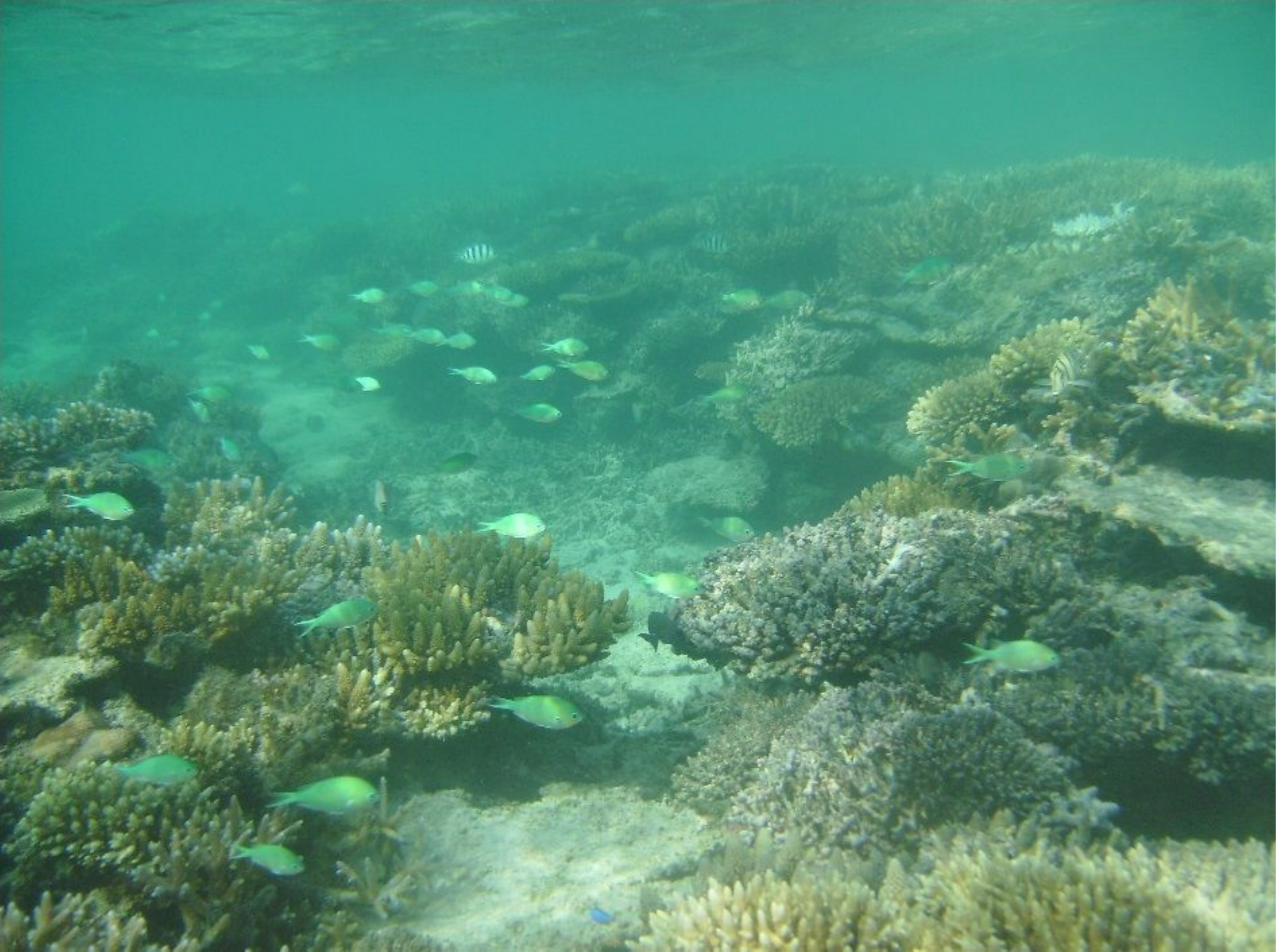} & \includegraphics[width=0.098\textwidth]{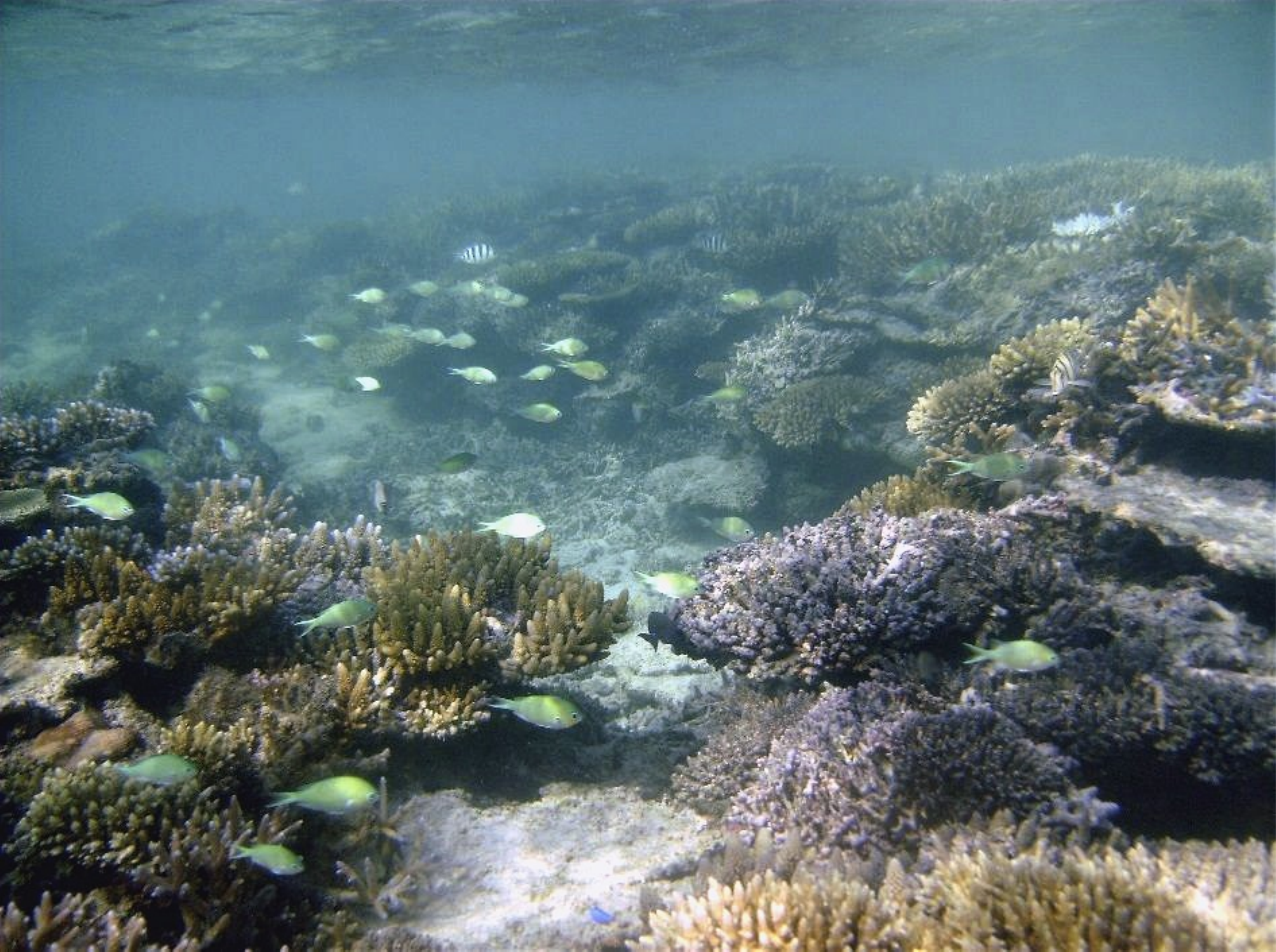} & \includegraphics[width=0.098\textwidth]{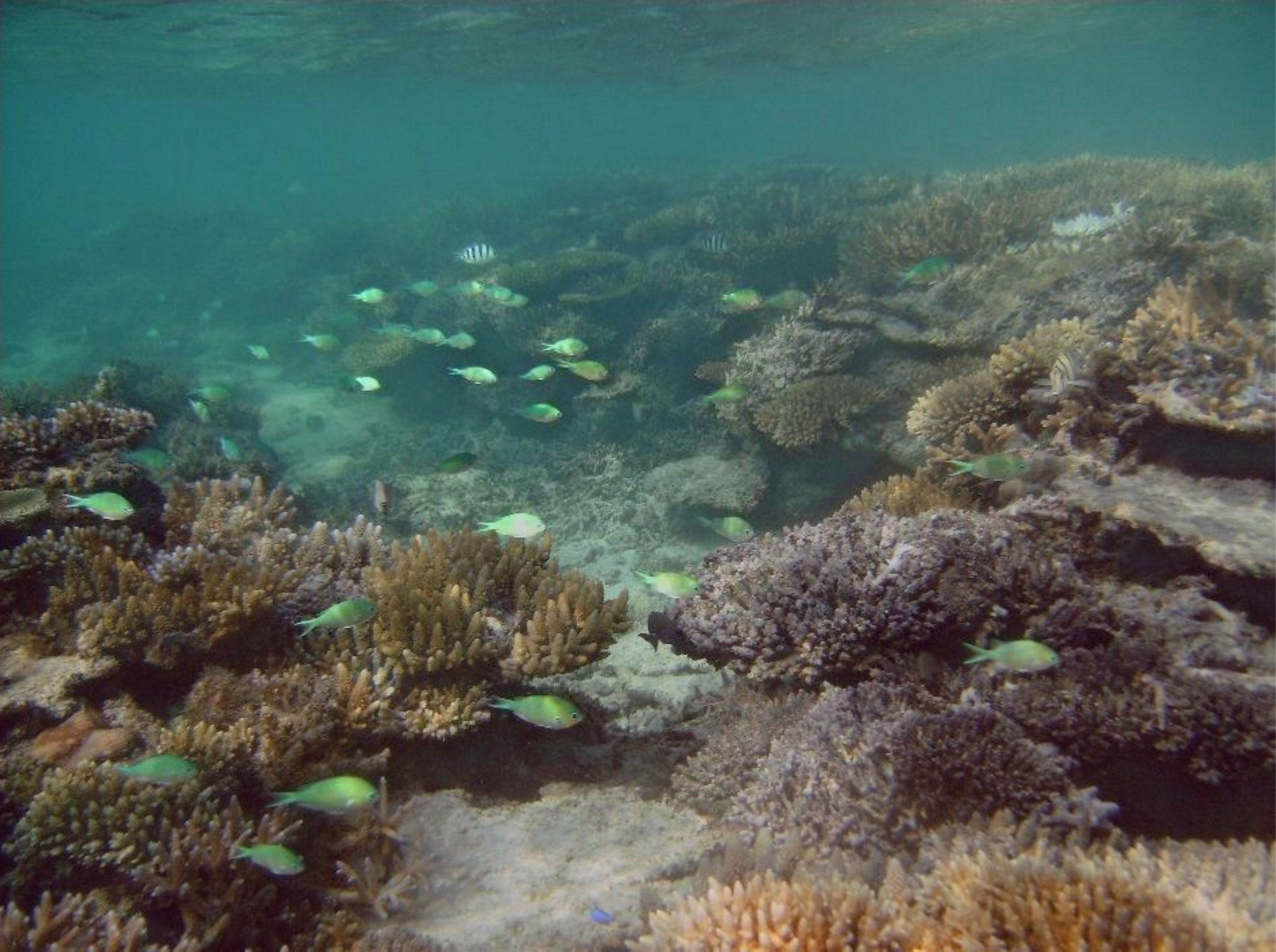} & \includegraphics[width=0.098\textwidth]{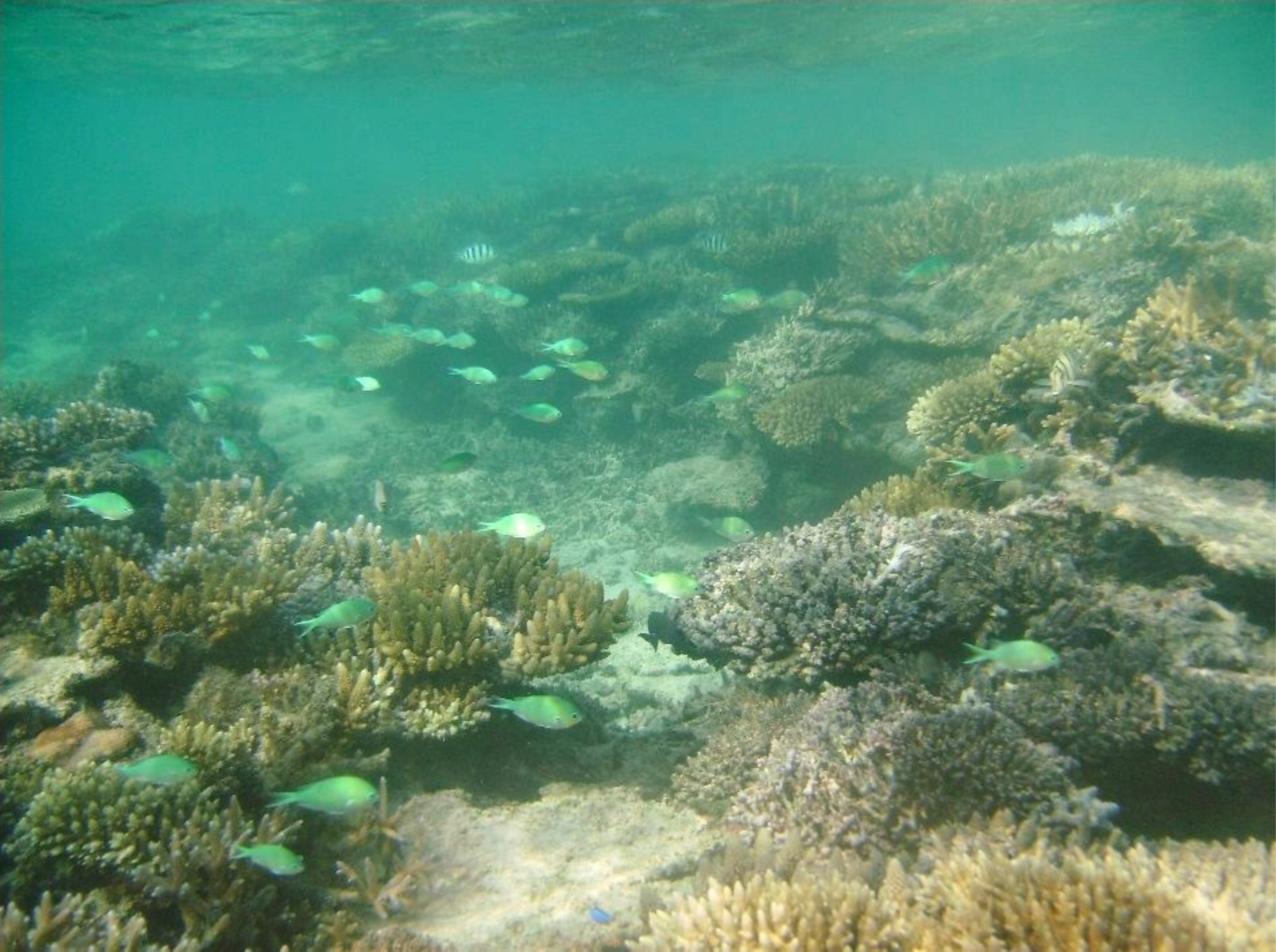} & \includegraphics[width=0.098\textwidth]{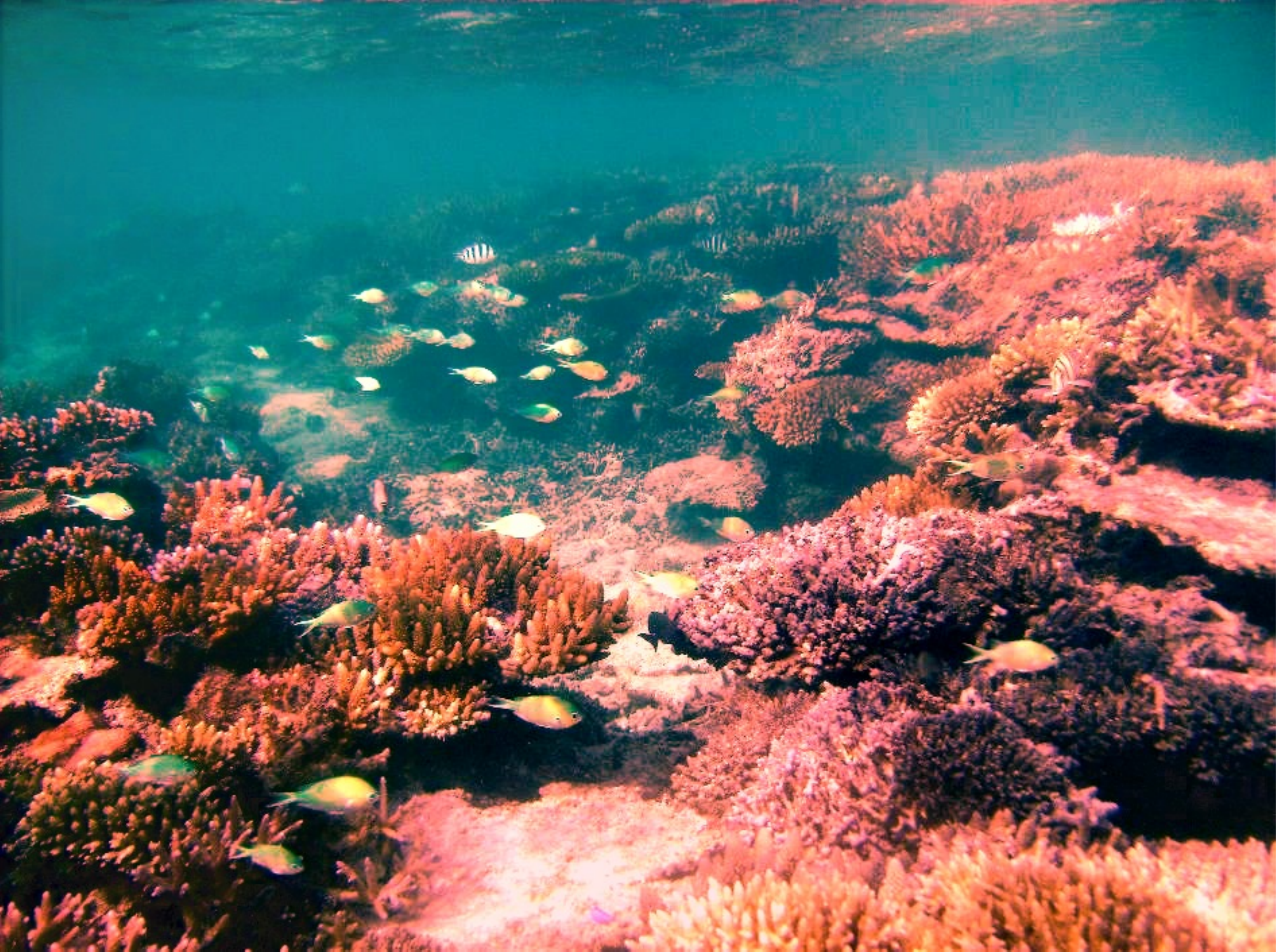} & \includegraphics[width=0.098\textwidth]{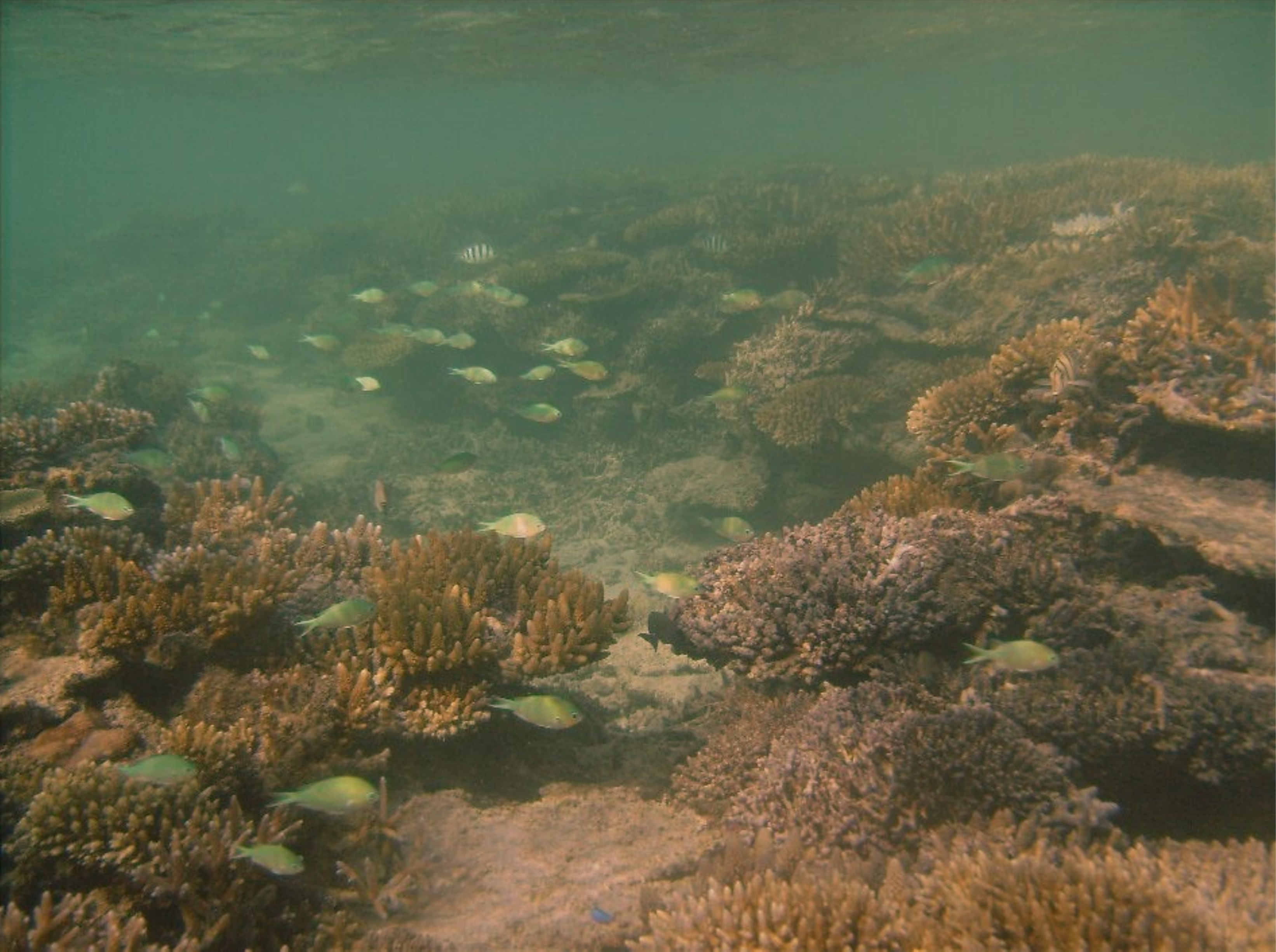}  & \includegraphics[width=0.098\textwidth]{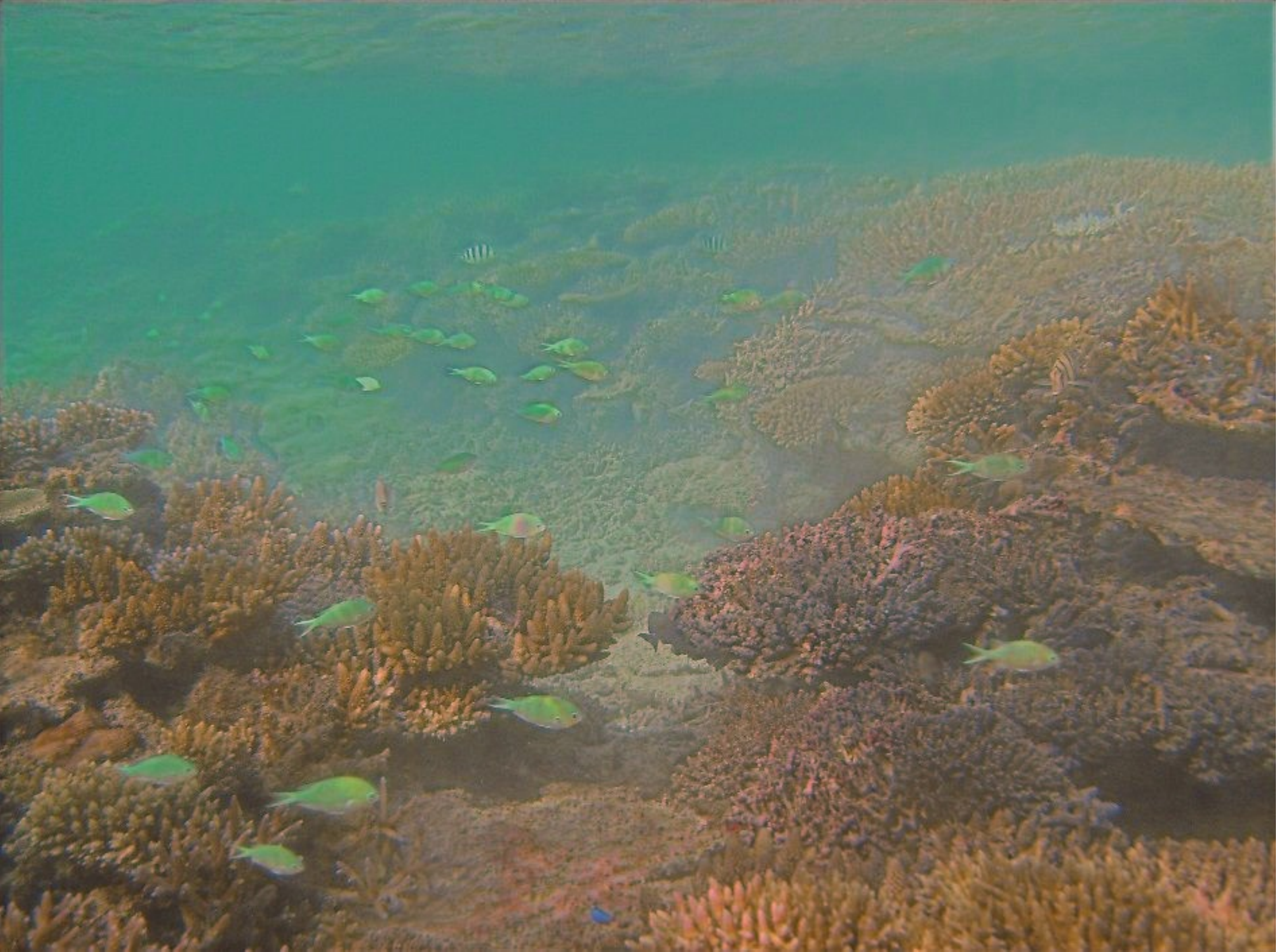} & \includegraphics[width=0.098\textwidth]{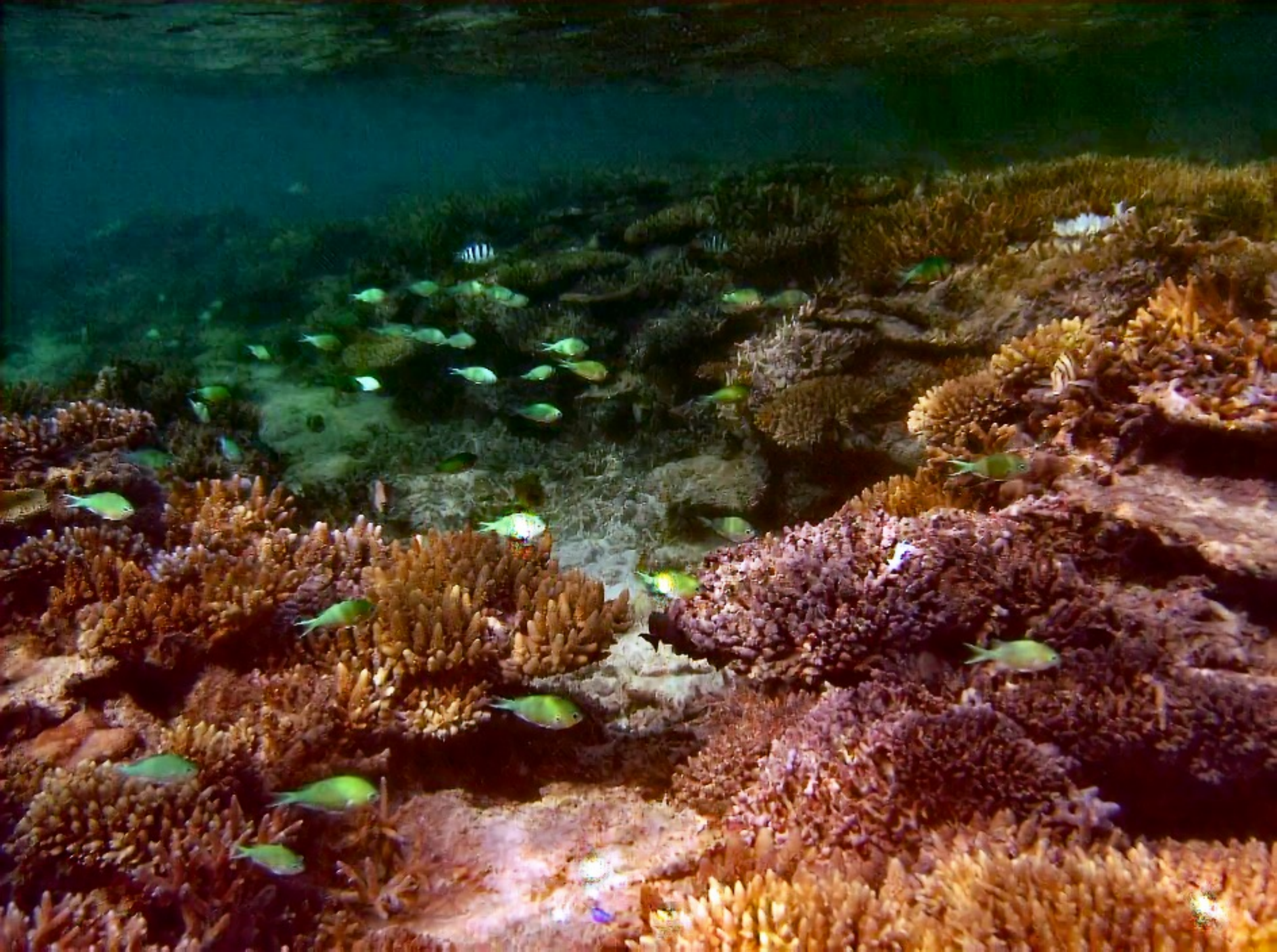} &\includegraphics[width=0.098\textwidth]{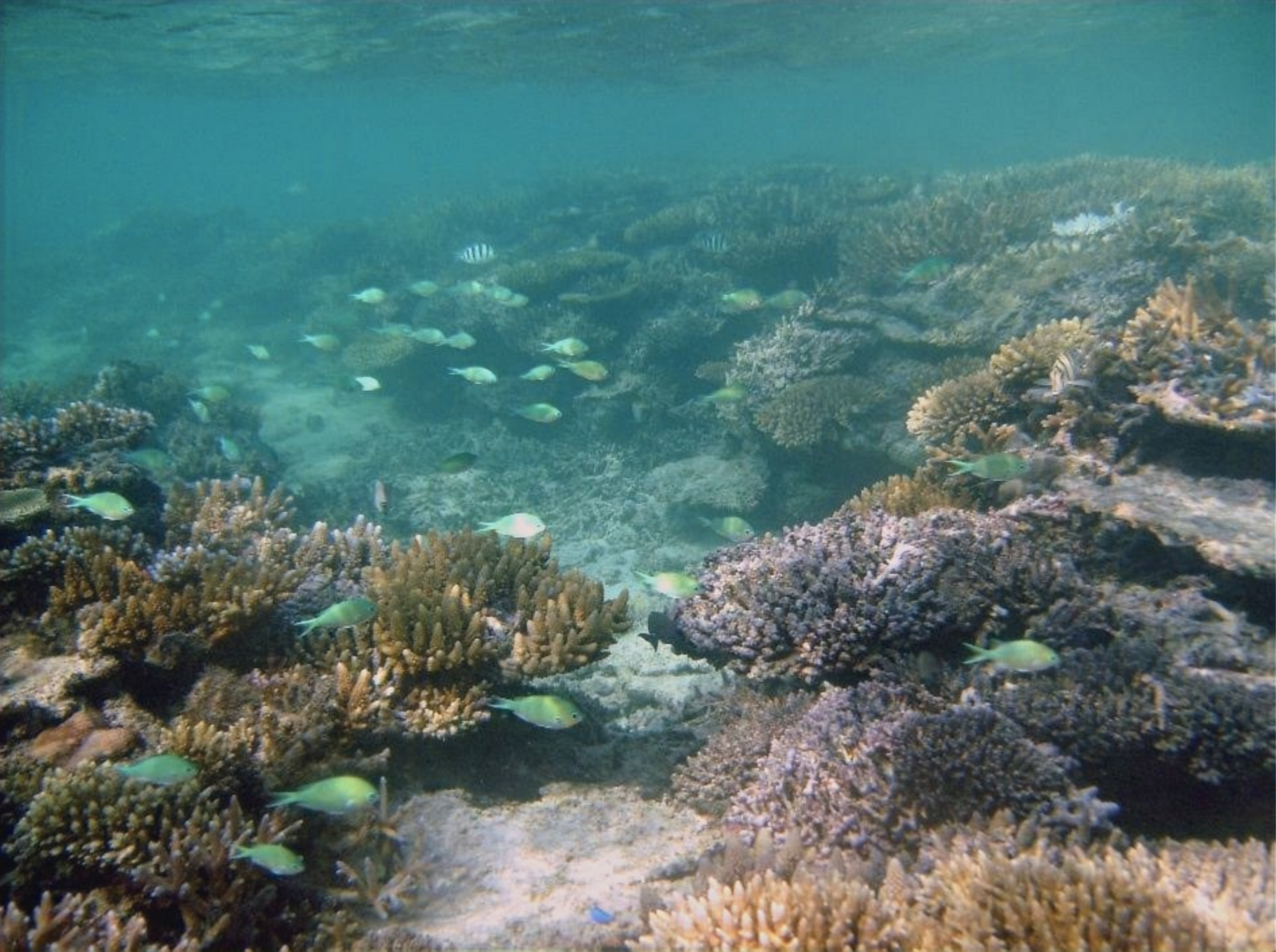} & \includegraphics[width=0.098\textwidth]{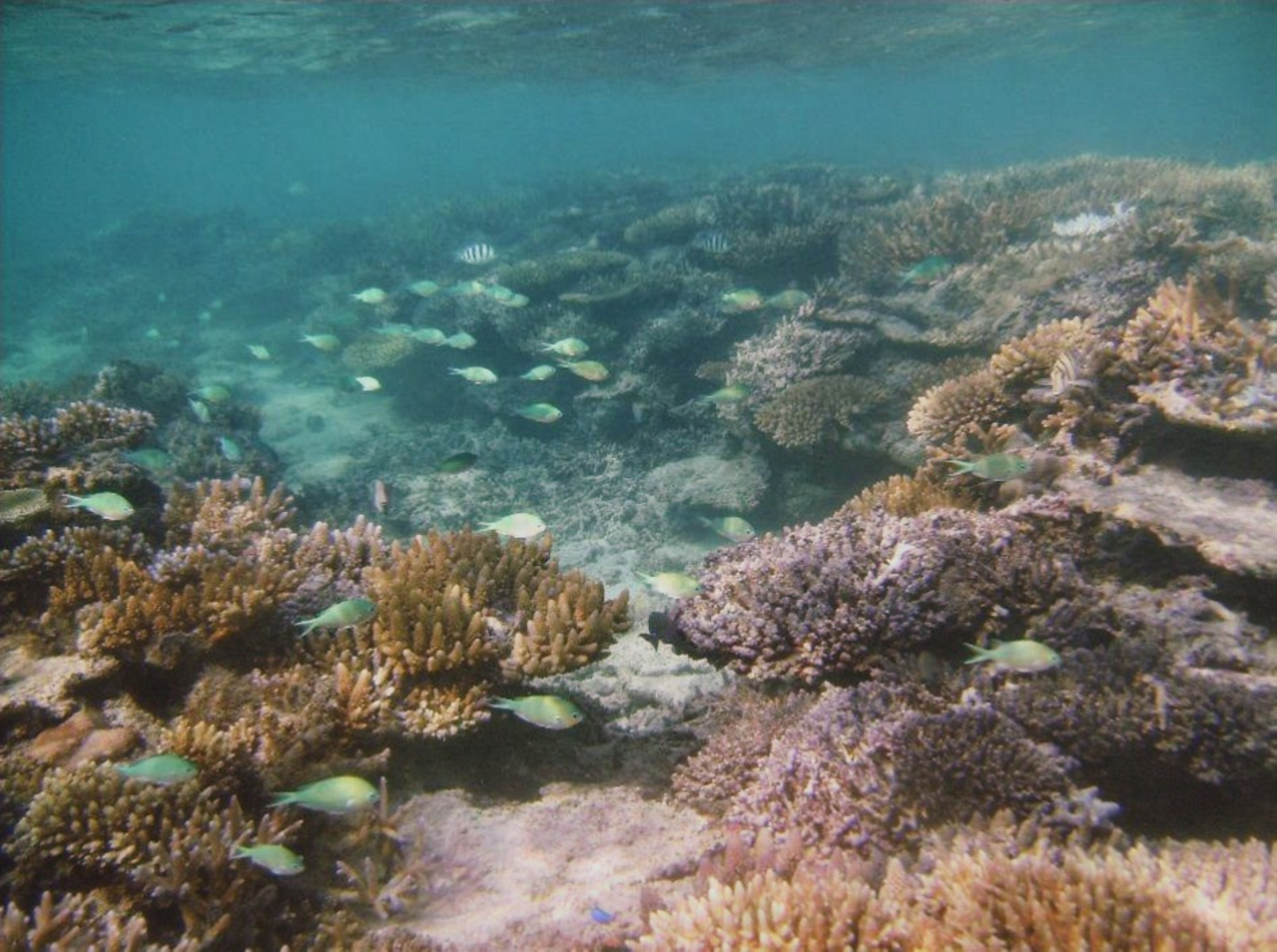} \\
Original & Fusion~\cite{Ancuti2018} & ARC~\cite{ARC} & DBL~\cite{BglightEstimationICIP2018_Submitted} & UWHL~\cite{BermanArvix_UWHL_dataset} & UWCNN-1~\cite{Anwar2018DeepUI_UWCNN}  & UWCNN-3~\cite{Anwar2018DeepUI_UWCNN}  & Cast-GAN~\cite{Li_2020_Cast-GAN} & WaterNET~\cite{WaterNET_TIP2019} & UColor~\cite{Ucolor}\\
\end{tabular}
\caption{Comparison of underwater image enhancement results for images that capture the two main water categories: oceanic  (rows 1-2) and coastal  (row 3).   More results are available at {https://puiqe.eecs.qmul.ac.uk}.}
\label{fig:: experiment_natural_images}
\end{figure*}

\section{Comparison}

We assess five measures on eight methods. We consider three underwater-specific IQA measures, namely UCIQE~\cite{UCIQE}, UIQM~\cite{UIQM_Panetta2016}, and CCF~\cite{WANG2018_NR_CFF}; and two colour accuracy measures, namely CIEDE2000~\cite{sharma2005ciede2000} and colour reproduction error~\cite{BMVCAngularError}. We compare their scores on eight enhancement methods: four physics-based methods, namely Automatic Red Channel (ARC)~\cite{ARC}, Depth-dependent Background Light (DBL)~\cite{BglightEstimationICIP2018_Submitted}, Fusion~\cite{Ancuti2018}, and Underwater Haze Line (UWHL)~\cite{BermanArvix_UWHL_dataset}; and four neural networks,  namely UWCNN~\cite{Anwar2018DeepUI_UWCNN}, Cast-GAN~\cite{Li_2020_Cast-GAN}, WaterNet~\cite{WaterNET_TIP2019} and UColor~\cite{Ucolor}. We show the results of UWCNN for the representative water type for oceanic (type I) and coastal water (type 3), denoted by UWCNN-I and UWCNN-3, respectively.

Fig.~\ref{fig:: experiment_natural_images} shows sample images commonly used in underwater literature. Table~\ref{Table:: IQAs_natural_images} reports the values of measures for nine methods applied on the sample images. Since the three IQA measures have different ranges, we use the original images' measured quality as a reference: smaller values than for the original image (red shade) indicate that the measure suggests that the enhancement methods improved image quality. It is possible to notice that the IQA measures fail to capture (perceptually) obvious distortions, such as unnatural red introduced to the water region by UWHL. However, all measures indicate UWHL has improved the image quality and deem 
the images enhanced by UWHL to be better than the original. 

Another important observation is that similar values do not refer to similar visual quality, for example, the UIQM values of the coral image (R2) enhanced by UWCNN-3 and Ucolor only differ by 0.01 (1.28 and 1.29, respectively) but the image enhanced by UWCNN-3 has a washout background and unnaturally red reef structure that are visually different from the UColor image. This shows that the numerical values do not correspond to changes in the perceived quality. Moreover, none of the measures has a valid range of values for comparison and it is unclear what a difference in the measure's value corresponds to the perceptual quality. 

\begin{table}[t!]
\caption{Comparison of underwater-specific IQA measures. For all measures: the higher the value, the better the image quality. The best value for each image is in bold font. A red shade indicates that the value is smaller than the original (Ori.), hence worse quality as depicted by the measure. R$k$ refers to the image (Img.) in the $k^{\text{th}}$ row in Fig.~\ref{fig:: experiment_natural_images}. }
\centering
\scriptsize
\setlength\tabcolsep{0.05pt}
\vspace{5pt}
\begin{tabular}{c}
\begin{tabular}{L{1.3cm}|C{0.6cm}|R{0.6cm}|R{0.65cm}R{0.65cm}R{0.65cm}R{0.65cm}R{0.65cm}R{0.65cm}R{0.65cm}R{0.65cm}R{0.65cm}R{0.65cm}R{0.65cm}R{0.65cm}}
\multicolumn{1}{c|}{{\textbf{IQA}}} &  \multicolumn{1}{c|}{{\textbf{Img.}}} &  \multicolumn{1}{c|}{{\textbf{Ori.}}} 
& \multicolumn{1}{c}{\cite{Ancuti2018}} & \multicolumn{1}{c}{\cite{ARC}} & \multicolumn{1}{c}{\cite{BglightEstimationICIP2018_Submitted}} & \multicolumn{1}{c}{\cite{BermanArvix_UWHL_dataset}} & \multicolumn{1}{c}{\cite{Anwar2018DeepUI_UWCNN}}-I  & \multicolumn{1}{c}{\cite{Anwar2018DeepUI_UWCNN}}-3 & \multicolumn{1}{c}{\cite{Li_2020_Cast-GAN}} & \multicolumn{1}{c}{\cite{WaterNET_TIP2019}} & \multicolumn{1}{c}{\cite{Ucolor}} \\
\hline 
\multirow{3}{*}{\rotatebox{0}{\textbf{UCIQE~\cite{UCIQE}}}} &  R1 & 0.43 & {0.62} & {0.56} & {0.48} & 0.69 & \RedCell{0.42} & \RedCell{0.42}   & \textbf{0.70} & {0.58} & {0.54} \\
 & R2 & 0.54 & {0.65} & {0.61} & {0.59} & {\textbf{0.67}} & \RedCell{0.50}  & {0.54} & 0.54 & {0.63} & {0.60}  \\
 & R3 & 0.54 & {0.64} & {0.58} & {0.58} & {\textbf{0.67}} & \RedCell{0.54} & {0.55} & 0.66 & {0.62} & {0.59}
\\
 \hline
 \multirow{3}{*}{\rotatebox{0}{\textbf{UIQM~\cite{UIQM_Panetta2016}}}} & R1 & 0.64 & {1.21} & {1.09} & {0.81} & {\textbf{1.86}} & {0.71} & {0.69} & 1.62 & {1.38} & {1.03} \\
 & R2 & 1.12 & {1.41} & {1.31} & {1.26} & {{1.49}} & {1.21} & {1.28} & \textbf{2.81} &{1.36} & {1.29} \\
 & R3 & 1.29 & {1.50}  & {1.31} & {1.34} & {{1.57}} & \RedCell{1.24}   & \RedCell{1.26}   & \textbf{1.78} & {1.43} & {1.31}\\\hline
 \multirow{3}{*}{\rotatebox{0}{\textbf{CCF~\cite{WANG2018_NR_CFF}}}}  & R1 & 8.64  & {17.26} & {12.16} & {10.71} & {\textbf{60.29}} & \RedCell{7.71}  & \RedCell{7.48}  & 40.47&  {17.78} & {13.50}  \\
 & R2 & 15.91 & {23.69} & {17.77} & {19.22} & {\textbf{36.83}} & \RedCell{12.69} & \RedCell{12.12} & 28.82 & {21.14} & {20.70}  \\
 & R3 & 22.17 & \RedCell{21.65}   & \RedCell{16.34}   & {34.29} & {{37.72}} & \RedCell{14.18} & \RedCell{13.04} & \textbf{59.64} & {{48.83}} & \RedCell{18.41}  \\
\end{tabular}%
\end{tabular}
\label{Table:: IQAs_natural_images}
\vspace{-10pt}
\end{table}

\begin{figure*}[t!]
\setlength\tabcolsep{0.6pt}
\centering
\scriptsize
\vspace{10pt}
\begin{tabular}{ccccccccccccc}
 &
\includegraphics[width=0.0885\textwidth]{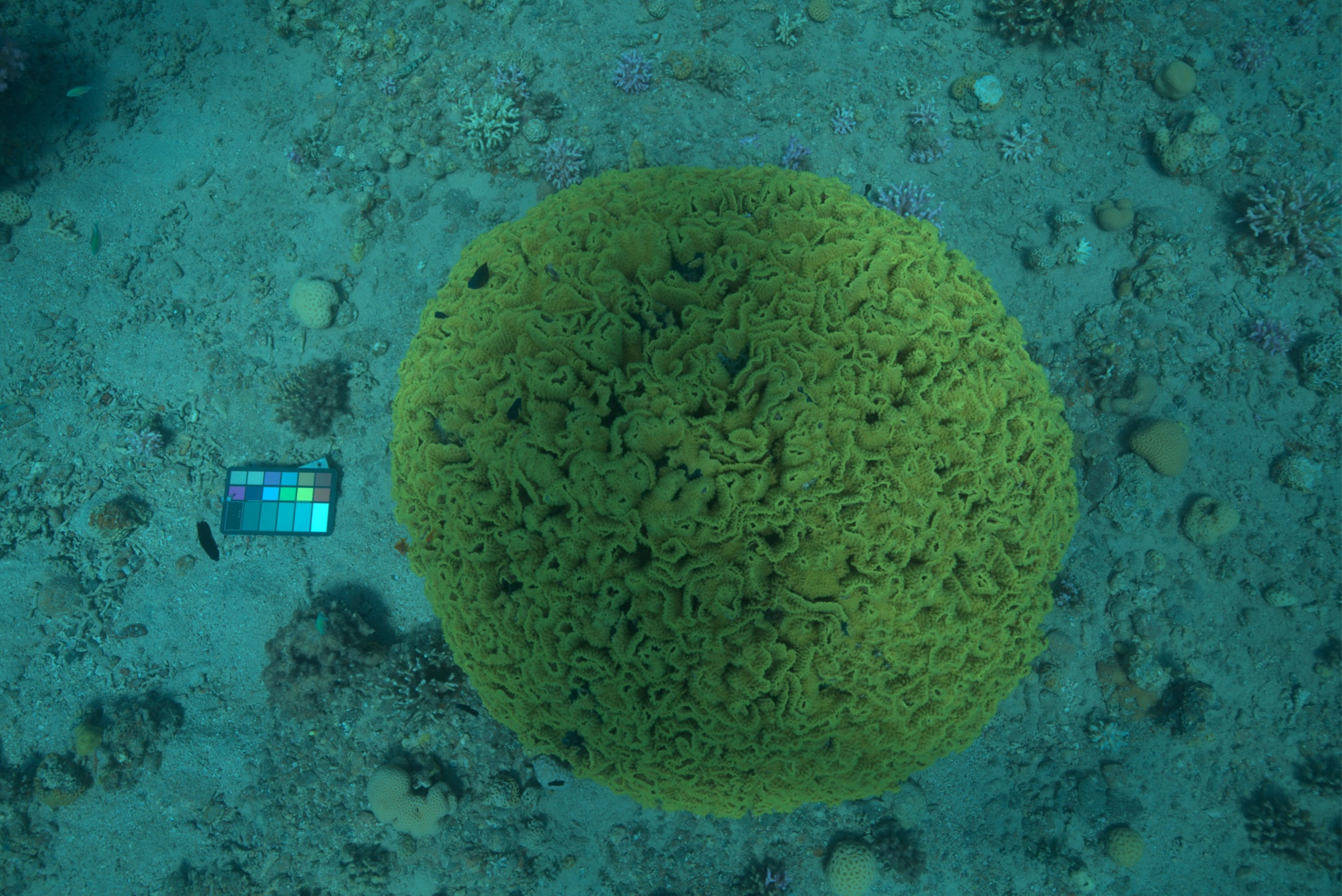} & \includegraphics[width=0.0885\textwidth]{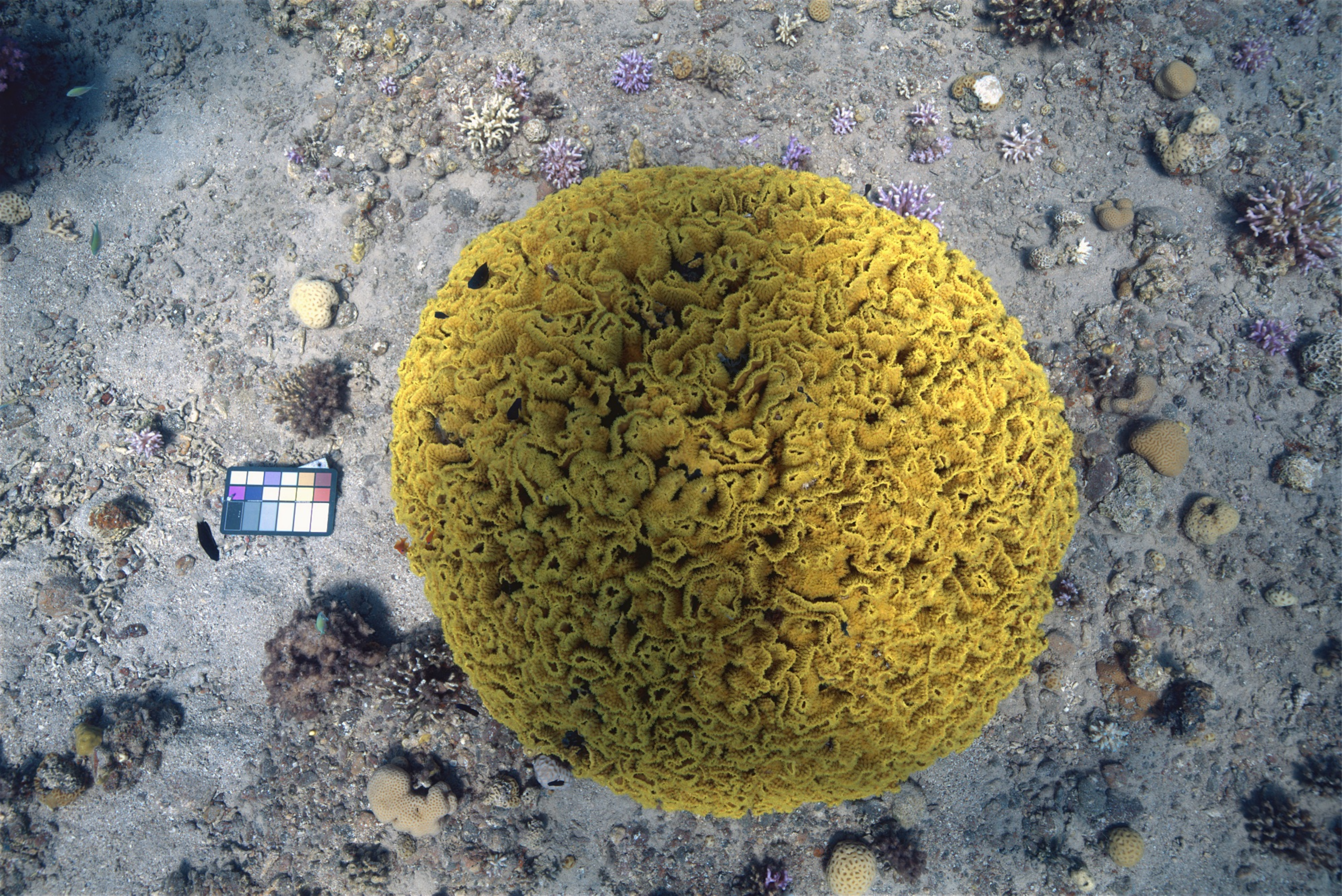} &  \includegraphics[width=0.0885\textwidth]{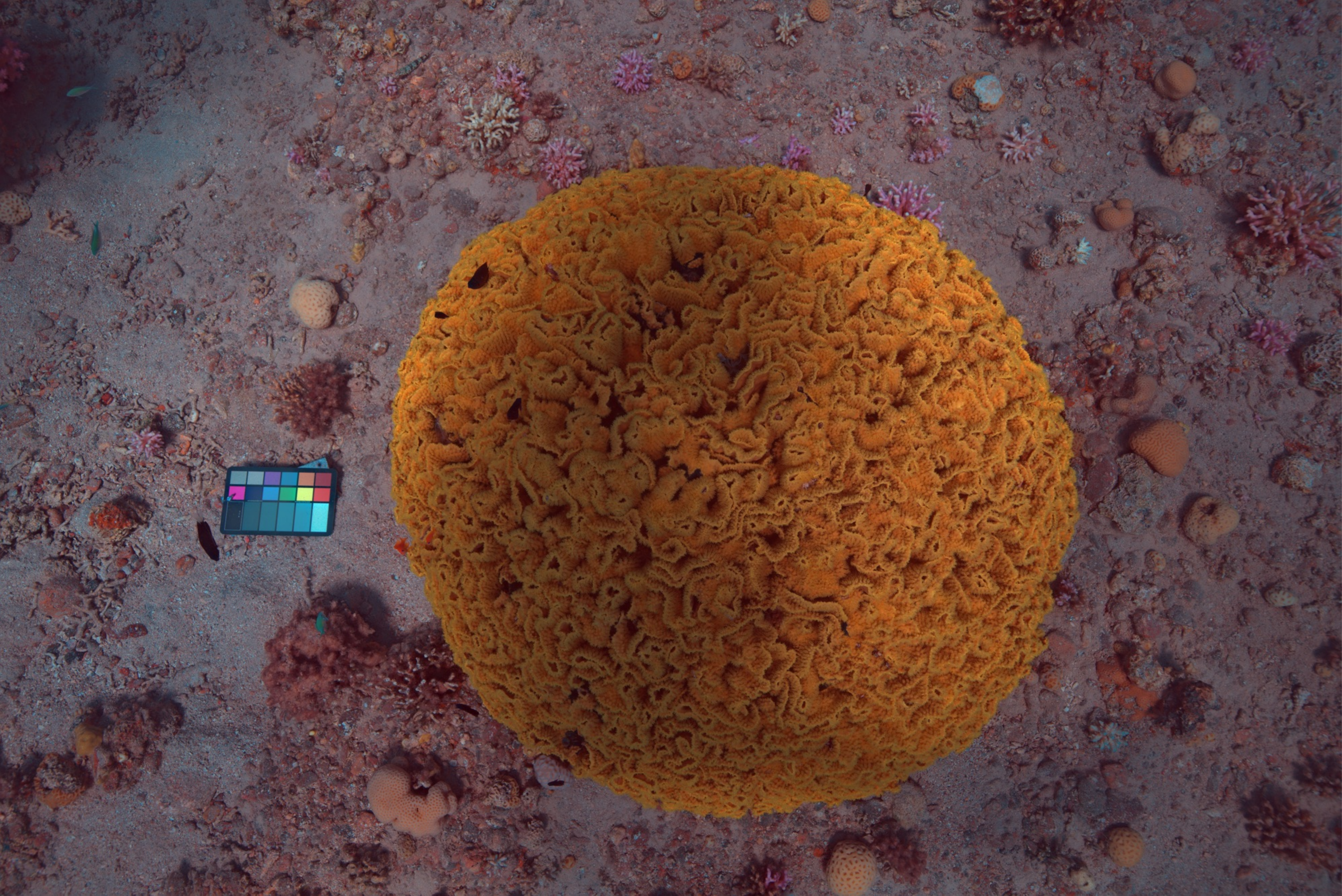} & \includegraphics[width=0.0885\textwidth]{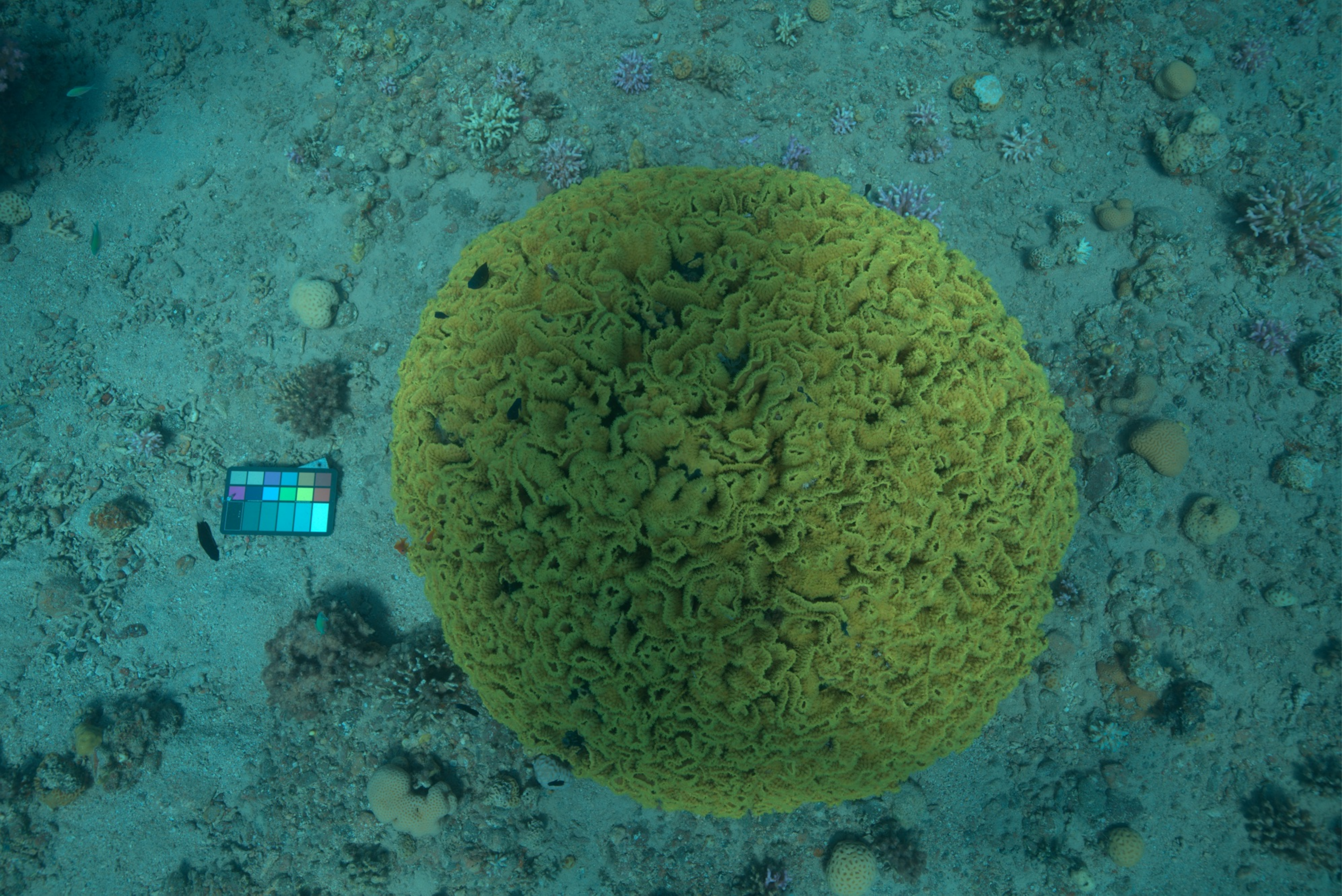}& \includegraphics[width=0.0885\textwidth]{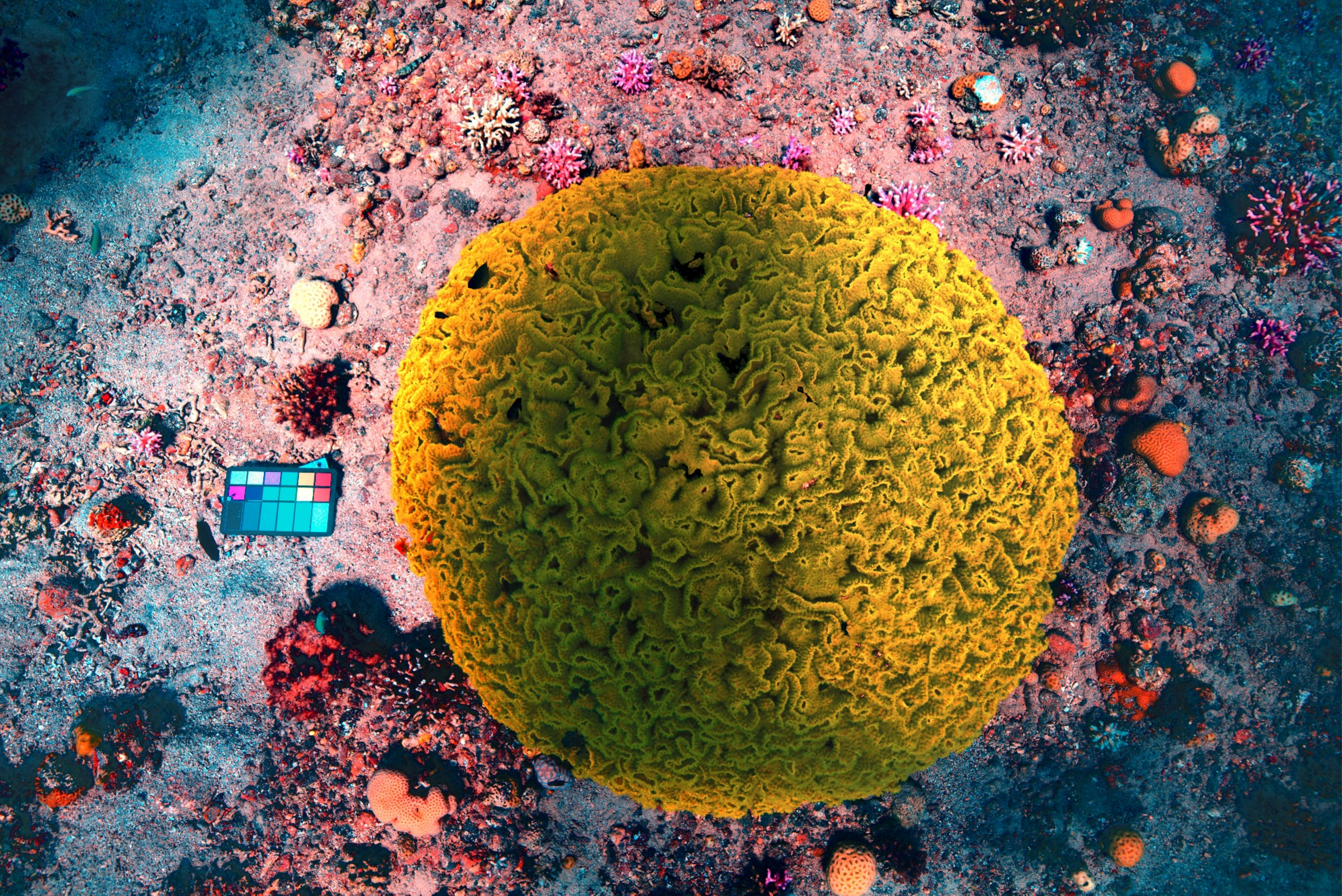}& \includegraphics[width=0.0885\textwidth]{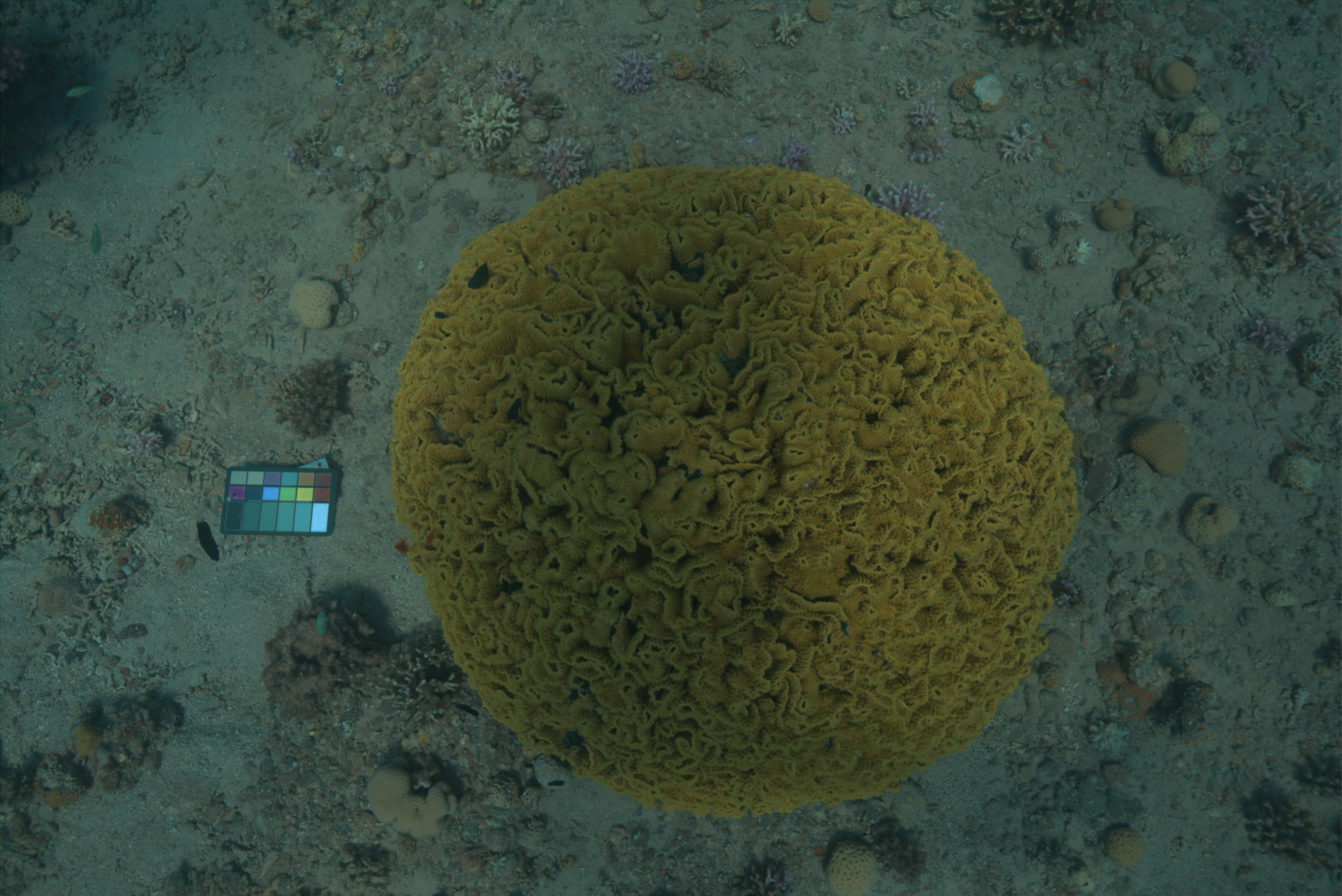} & \includegraphics[width=0.0885\textwidth]{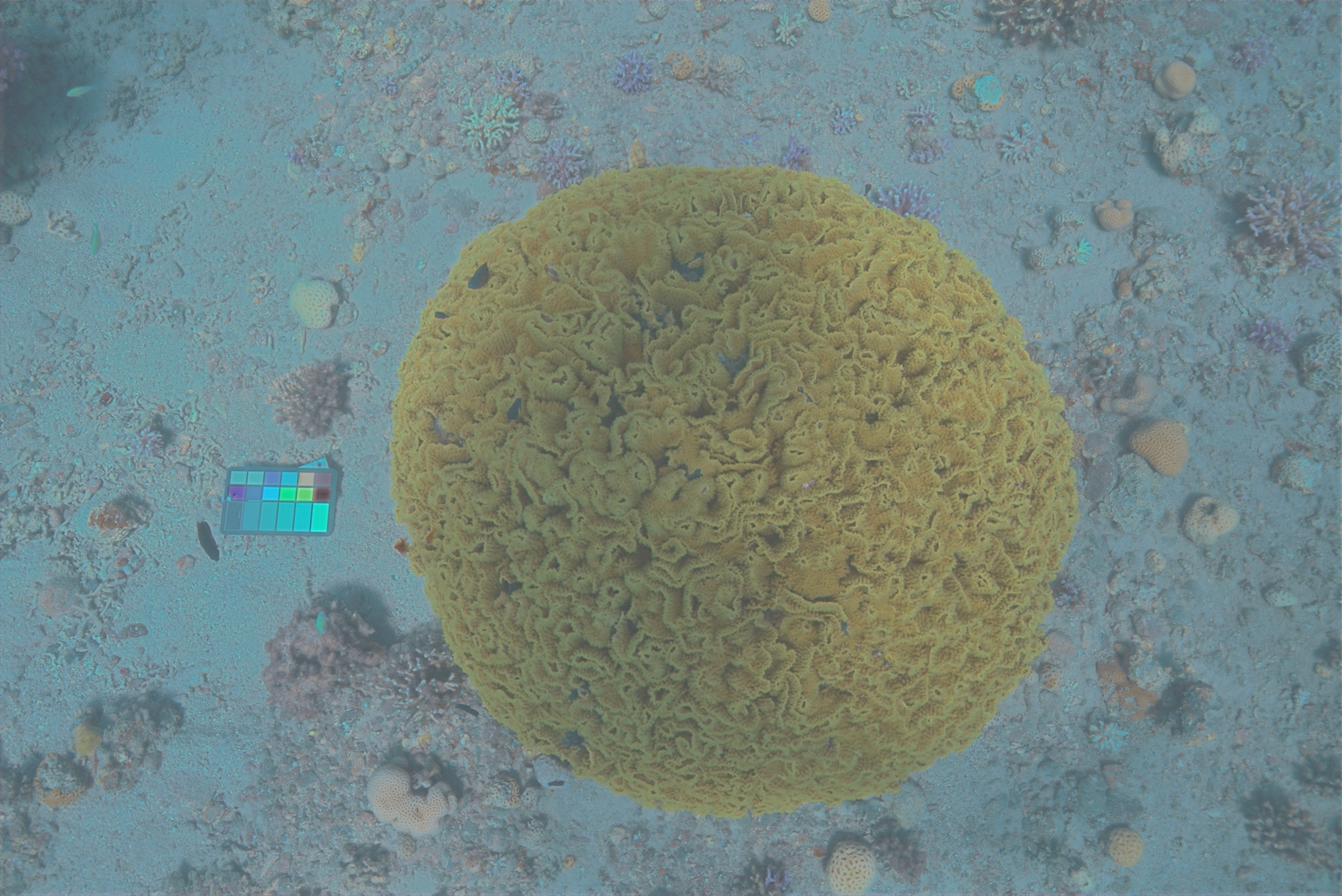}  & \includegraphics[width=0.0885\textwidth]{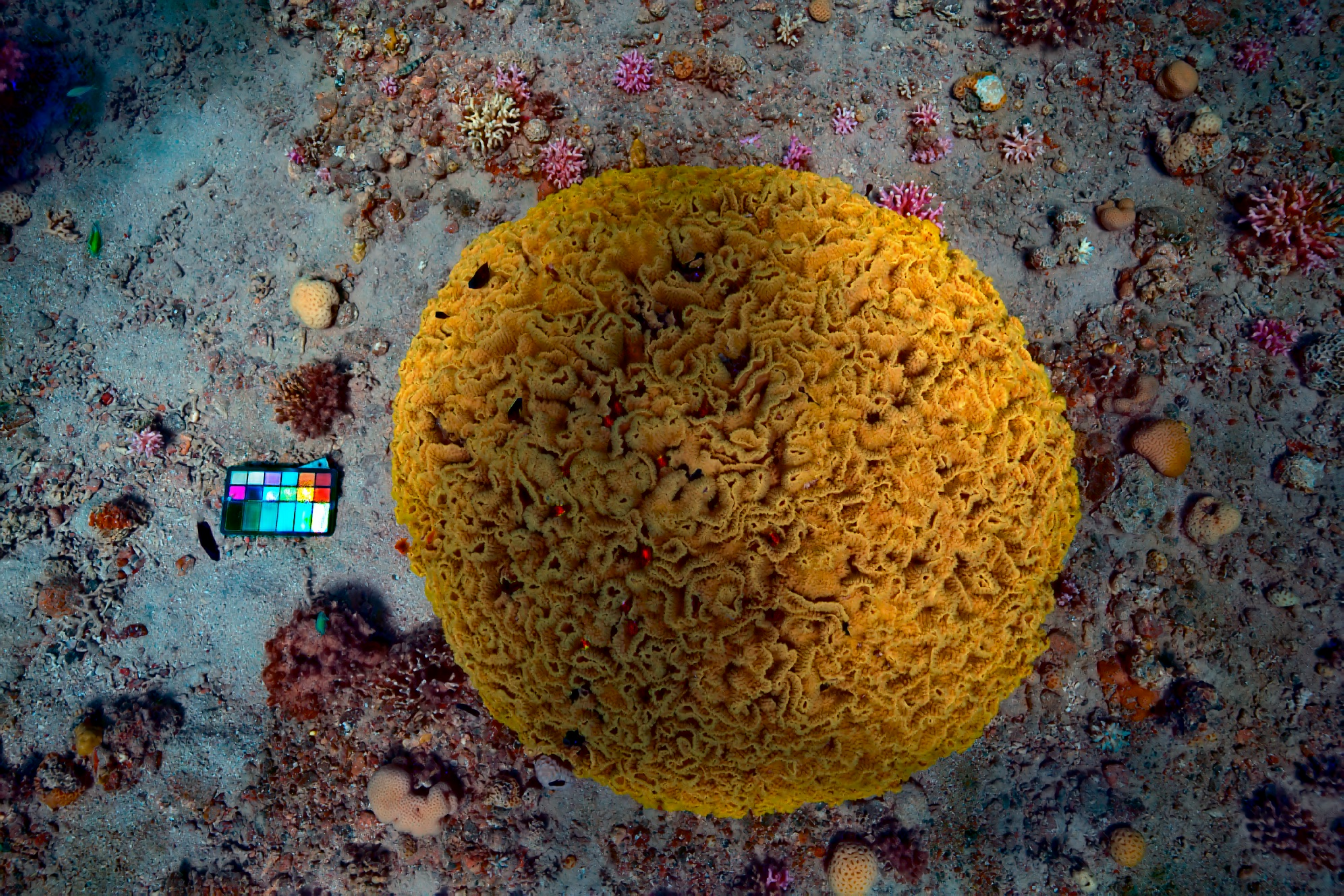}  &  \includegraphics[width=0.0885\textwidth]{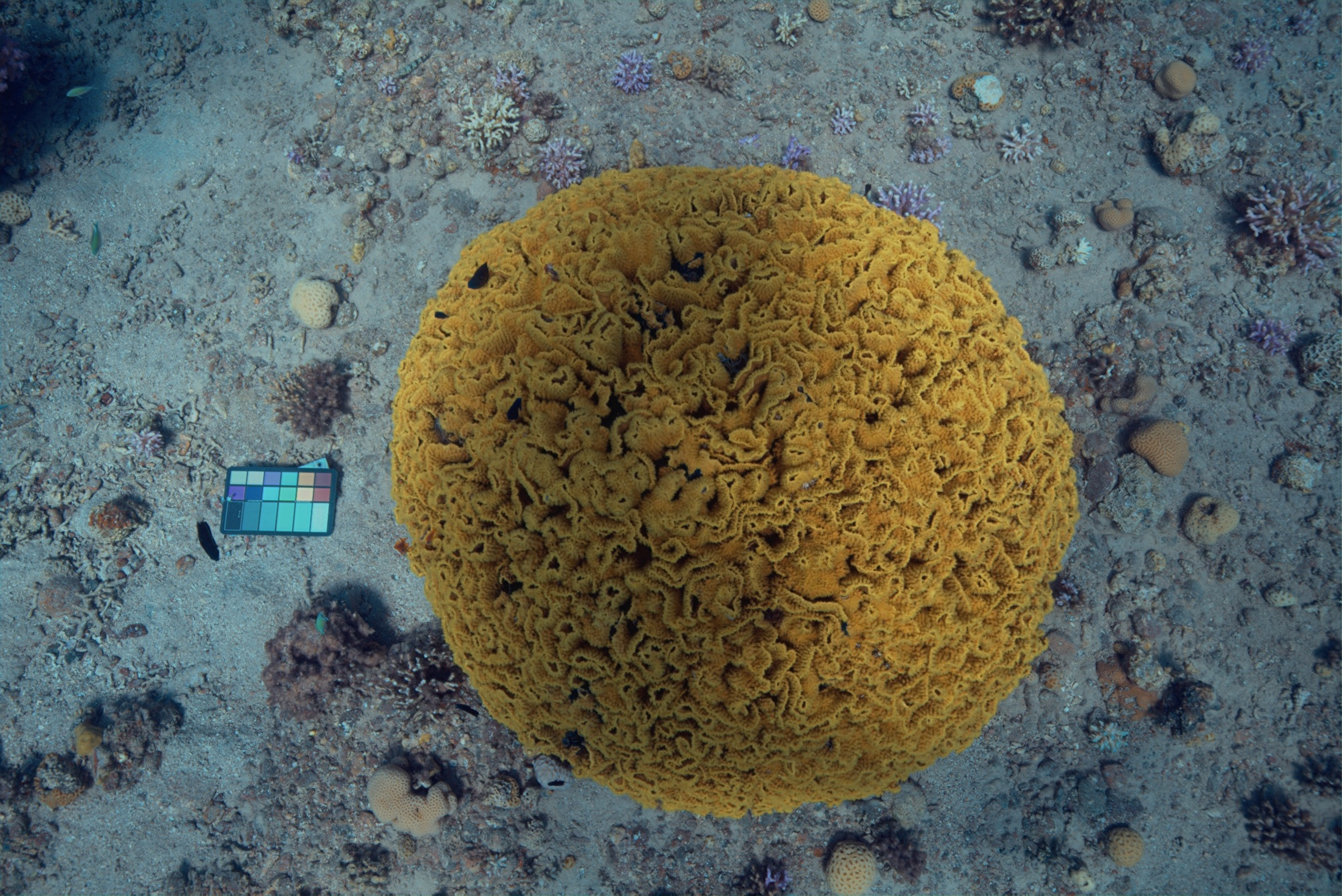} & \includegraphics[width=0.0885\textwidth]{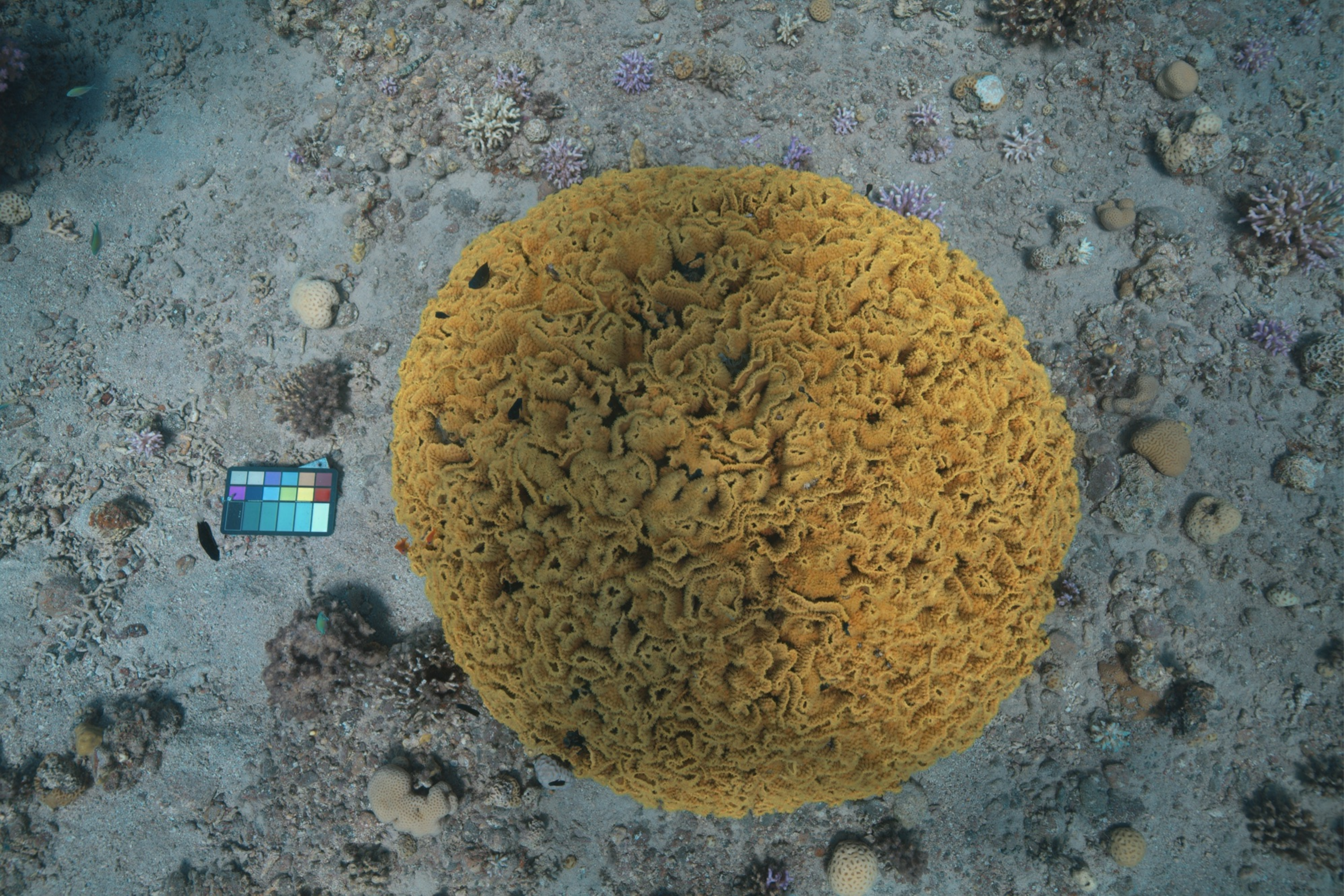}\\
{\includegraphics[width=0.085\textwidth]{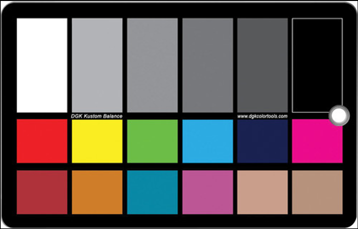} }&\includegraphics[width=0.0885\textwidth]{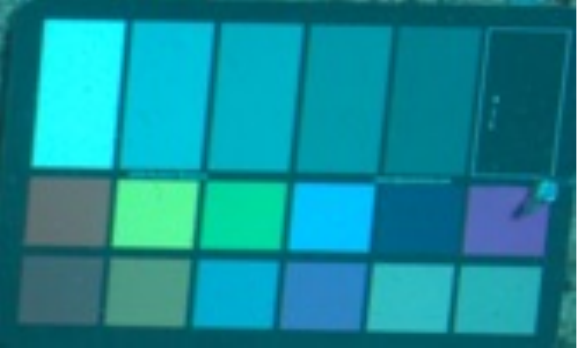} & \includegraphics[width=0.0885\textwidth]{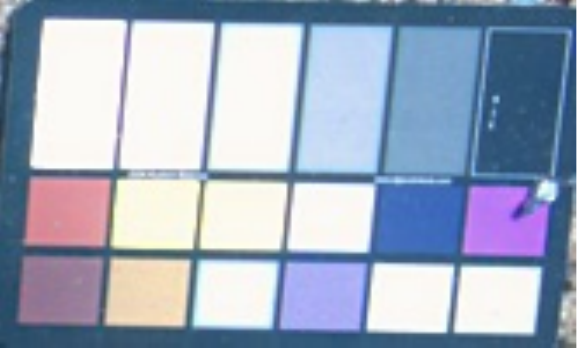} &  \includegraphics[width=0.0885\textwidth]{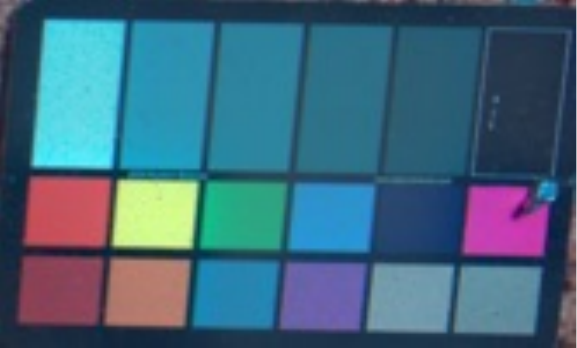} & \includegraphics[width=0.0885\textwidth]{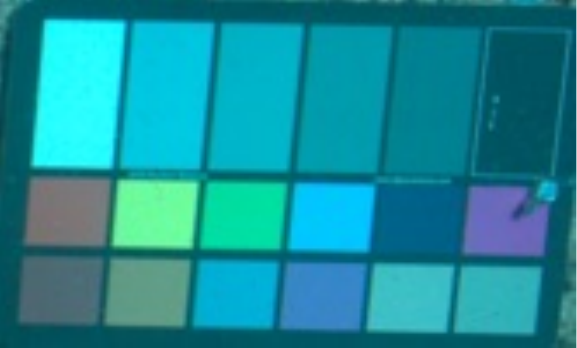}& \includegraphics[width=0.0885\textwidth]{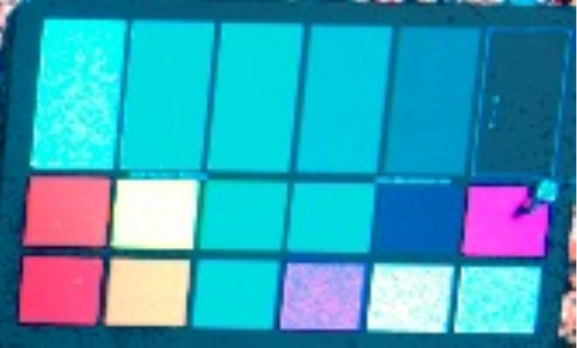}& \includegraphics[width=0.0885\textwidth]{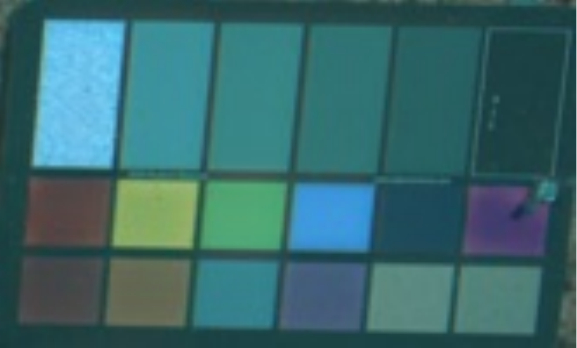} & \includegraphics[width=0.0885\textwidth]{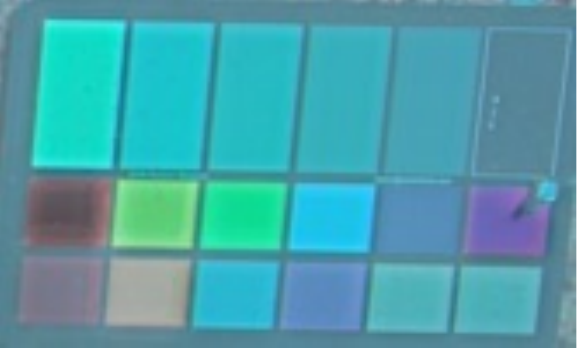}  & \includegraphics[width=0.0885\textwidth]{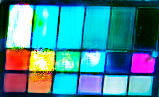} & \includegraphics[width=0.0885\textwidth]{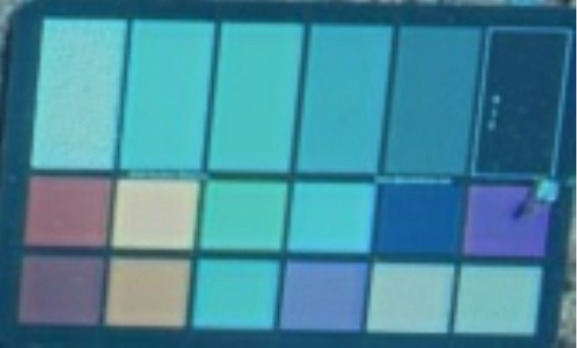} & \includegraphics[width=0.0885\textwidth]{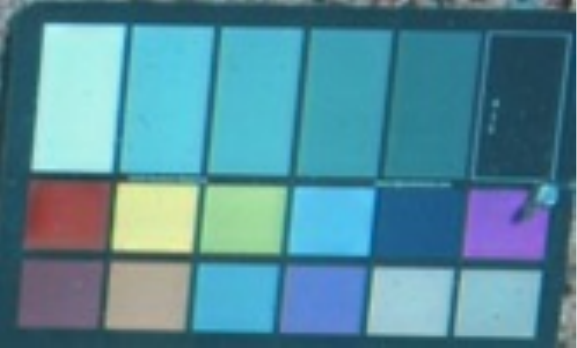}\\
\scriptsize{Reference} &\scriptsize{Original} & \scriptsize{Fusion~\cite{Ancuti2018}} & \scriptsize{ARC~\cite{ARC}} & \scriptsize{DBL~\cite{BglightEstimationICIP2018_Submitted}} & \scriptsize{UWHL~\cite{BermanArvix_UWHL_dataset}} & \scriptsize{UWCNN-1~\cite{Anwar2018DeepUI_UWCNN}}  & \scriptsize{UWCNN-3~\cite{Anwar2018DeepUI_UWCNN}}  & \scriptsize{Cast-GAN~\cite{Li_2020_Cast-GAN}} & \scriptsize{WaterNET~\cite{WaterNET_TIP2019}} & \scriptsize{UColor~\cite{Ucolor}}\\
\end{tabular}
\caption{Colour accuracy is measured using images with a reference colour checker. The left-most column shows the colour checker under a reference illuminant (D65). The remaining columns show the colour checker as captured underwater (Sea-thru dataset~\cite{seathru2019CVPR}) and sample enhanced images, with corresponding enlarged crop of the colour checker in the bottom row.
}
\label{fig:: error_at_range_checker}
\end{figure*}

 Fig.~\ref{fig:: error_at_range_checker} shows the reference colour checker and sample enhanced images from the Sea-thru dataset. Since the dataset does not provide any annotations,
 {we manually segmented and labelled the colour patches using~\mbox{LabelMe}~\cite{LabelMe}. The raw images were converted to 8-bit PNG using the macOS Preview export function and were resized with bilinear interpolation to one-fourth of the original size.} The six achromatic patches on the reference colour checker include four grey patches with different lightness. Note that none of the methods could enhancement the colour checker to a value close to that of the reference chart: for example UWCNN-3 de-saturated the colour and Fusion turned two grey patches into white patches. 
 
\begin{table}[t!]
\caption{Colour accuracy measures: the reproduction angular error ($\Phi$)~\cite{BMVCAngularError} and CIEDE2000~($\Delta E_{00}$)~\cite{sharma2005ciede2000} of the 6 achromatic patches in the colour checker. Note that CIEDE2000 is computed with respect to the reference colour checker. For all measures: the smaller the error, the higher the colour accuracy, hence the better the colour correction. The smallest error value of each patch is bold. A red shade indicates that the error is larger than the original image (Ori.).}
\vspace{10pt}
\scriptsize
\setlength\tabcolsep{0.1pt}
\centering
\begin{tabular}{c}
\begin{tabular}{C{0.7cm}|C{0.7cm}|R{0.65cm}|R{0.70cm}R{0.70cm}R{0.70cm}R{0.70cm}R{0.70cm}R{0.70cm}R{0.70cm}R{0.70cm}R{0.70cm}R{0.70cm}R{0.70cm}R{0.70cm}}
\multicolumn{1}{c|}{\textbf{Mea.}} & \multicolumn{1}{c|}{\textbf{Patch}} &  \multicolumn{1}{c|}{\textbf{Ori.}} & \multicolumn{1}{c}{\cite{Ancuti2018}} & \multicolumn{1}{c}{\cite{ARC}} & \multicolumn{1}{c}{\cite{BglightEstimationICIP2018_Submitted}} & \multicolumn{1}{c}{\cite{BermanArvix_UWHL_dataset}} & \multicolumn{1}{c}{\cite{Anwar2018DeepUI_UWCNN}-I}  & \multicolumn{1}{c}{\cite{Anwar2018DeepUI_UWCNN}-3} & \multicolumn{1}{c}{\cite{Li_2020_Cast-GAN}} &  \multicolumn{1}{c}{\cite{WaterNET_TIP2019}} & \multicolumn{1}{c}{\cite{Ucolor}} \\
\hline 
 \multirow{6}{*}{$\Phi$}& \cellcolor{patch_13} & 30.69 & {\textbf{0.23}}& {20.84}    & {29.78}    & {28.73}    & {16.87}    & {30.54}    & 2.67 
& {13.15}    & {9.14}  \\
&  \cellcolor{patch_14} & 35.39 & {\textbf{0.16}}& {25.88}    & \RedCell{35.41} & {35.29}    & {19.40}& {28.92}  & \RedCell{32.19}  & {20.11}    & {19.03} \\
&  \cellcolor{patch_15} & 35.37 & {\textbf{0.39}}& {23.16}    & \RedCell{35.39} & {35.27}    & {18.68}    & {26.28}  &
\RedCell{35.29 }
  & {20.20}& {19.94} \\
&  \cellcolor{patch_16} & 35.38 & {\textbf{6.22}}& {21.36}    & \RedCell{35.39} & \RedCell{35.38} & {18.91}    & {22.6}& \RedCell{35.34}
 & {22.20}& {21.36} \\
&  \cellcolor{patch_17} & 35.29 & {\textbf{11.58}}    & {20.43}    & \RedCell{35.31} & \RedCell{35.89} & {19.38}    & {18.81}   & \RedCell{35.30}  & {23.84}    & {22.15} \\
&  \cellcolor{patch_18} & 34.35 & {18.08}    & {16.56}    & \RedCell{34.54} & {32.86}    & {18.29}    & {\textbf{12.76}}  & \RedCell{38.79}  & {25.84}    & {19.36} \\ \hline
 \multirow{6}{*}{$\Delta E_{00}$} & \cellcolor{patch_13} & 26.06 & {\textbf{3.64}}& \RedCell{27.41} & {25.98}    & {25.83}    & \RedCell{27.74} & 13.37 & \RedCell{29.01} & {25.94}    & {20.76} \\
&  \cellcolor{patch_14} & 23.80  & {\textbf{13.98}}    & \RedCell{25.03} & \RedCell{23.82} & \RedCell{24.91} & \RedCell{25.55} & \RedCell{24.04} & 28.16 & {22.85}    & {21.71} \\
&  \cellcolor{patch_15} & 23.16 & {22.73}    & {21.62}    & \RedCell{23.37} & \RedCell{27.42} & {22.69}    & \RedCell{24.06} & \RedCell{27.59} & \RedCell{23.70}  & {\textbf{22.31}} \\
&  \cellcolor{patch_16} & 22.80  & \RedCell{27.85} & {\textbf{17.90}}& \RedCell{23.13} & \RedCell{30.42} & {19.91}    & \RedCell{25.88} & \RedCell{24.81} & \RedCell{23.89} & {22.40}  \\
&  \cellcolor{patch_17} & 21.90  & \RedCell{30.36} & {\textbf{14.35}}    & \RedCell{22.27} & \RedCell{28.46} & {16.79}    & \RedCell{30.66} & 19.21 & \RedCell{22.21} & {20.63} \\
&  \cellcolor{patch_18} & 28.57 & \RedCell{30.25} & {\textbf{18.82}}    & \RedCell{28.74} & \RedCell{31.1}  & {22.02}    & \RedCell{43.30} & 10.00 & {26.55}    & {24.02}
\end{tabular}%
\end{tabular}
\label{table: angular_error_6_patch}
\end{table}

\begin{figure}[t!]
\centering
\scriptsize
\vspace{15pt}
\begin{tabular}{ccccccccccccc}
\includegraphics[width=0.38\textwidth]{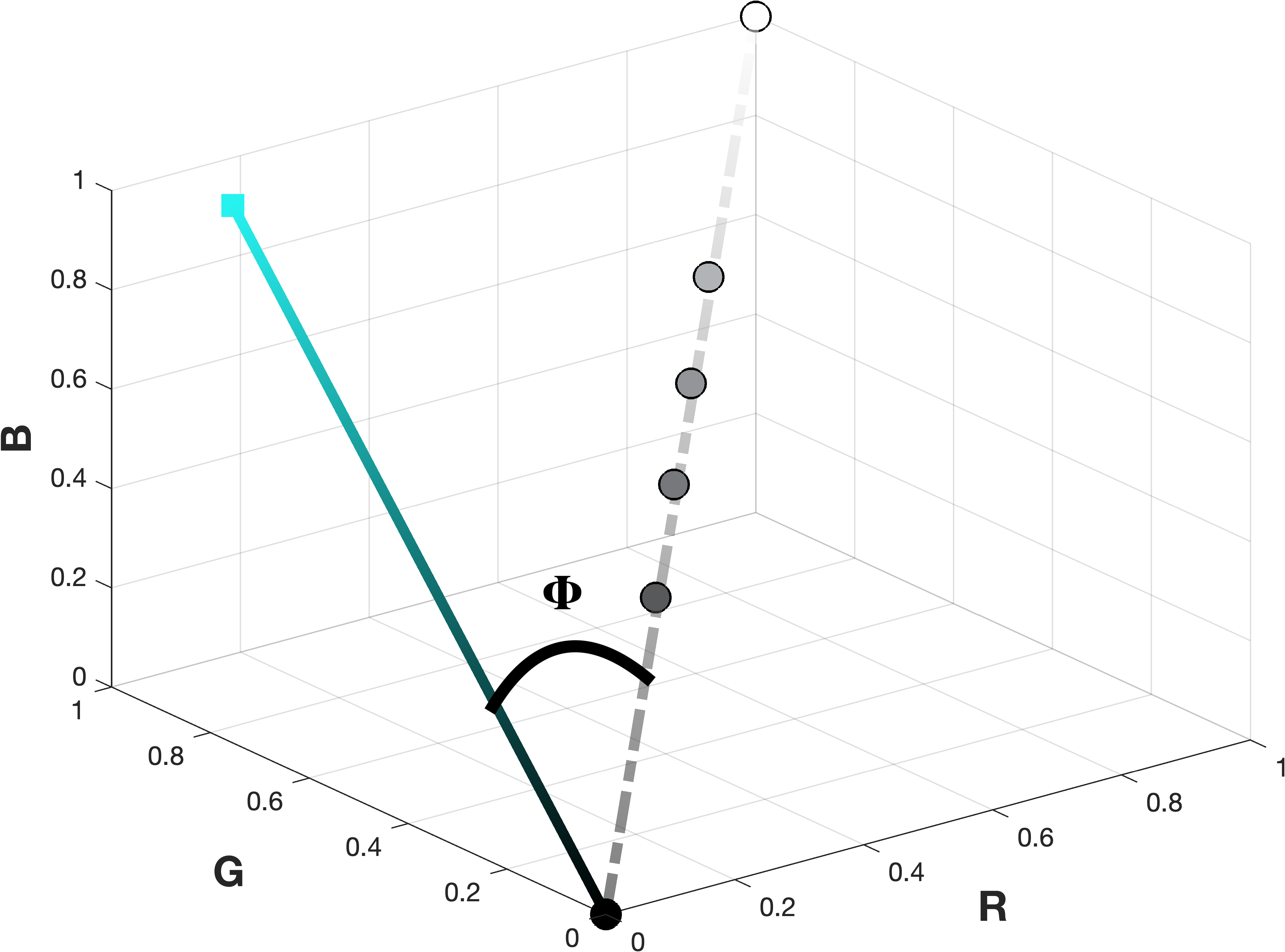} \\ 
\end{tabular}
\caption{The colour reproduction error measures the angle~$\Phi$ between the achromatic vector (diagonal) and the vector of the ‘corrected' patch in the RGB colour space. Any point on the ‘corrected' patch vector has the same error. (Circle markers: achromatic patches under reference illuminant; Square marker:  white patch in the image enhanced by UWHL.)}
\label{fig:: Phi_illustration}
\end{figure}

Table~\ref{table: angular_error_6_patch} reports the colour reproduction error,~$\Phi$, and CIEDE2000,~$\Delta E_{00}$. The reproduction error suggests that the best ‘colour accuracy' is achieved by Fusion with near-zero angular error for three of the patches. However,~$\Phi$ misses that  two patches were expected to be grey instead of white.  Moreover,~$\Phi$ does not account for the deterioration introduced by UWCNN-3 to the colour checker. To illustrate why~$\Phi$ fails to account for the colour distortion, we illustrate the measured error in Fig.~\ref{fig:: Phi_illustration}, using the colour of the white patch in the UWHL-enhanced image as an example. As~$\Phi$ only measures the angular error, any point on the corrected colour vector will have the same error. Hence the measure is unable to distinguish the perceived colour difference. Table~\ref{table: angular_error_6_patch} also shows that~$\Delta E_{00}$ is preferable to~$\Phi$ for measuring colour accuracy. In fact,~$\Delta E_{00}$ suggests that the colour difference of the grey patches in the Fusion-enhanced image can be perceived by humans (value exceeding 1) and~$\Delta E_{00}$ also correctly identifies the deterioration in colour accuracy by UWCNN-3. Moreover,~$\Delta E_{00}$  can be applied to quantify the colour difference between all of the 18 colour patches, ensuring a more holistic analysis of colour accuracy.

\section{Conclusion}
We discussed and evaluated the performance of quality measures commonly used for enhanced underwater images, including image quality measures specifically defined for the water medium and generic colour accuracy measures.  We showed that these image quality measures are unable to account for artefacts or changes in perceptual quality. For measuring colour accuracy, we showed that CIEDE2000 is preferable to the use of the colour reproduction error. Subjective tests with well-trained participants and colour accuracy using CIEDE2000 should be used in evaluations, until a suitable measure is developed and validated.  

\bibliographystyle{IEEEbib}
\bibliography{egbib.bib}
\end{document}